\documentclass[lettersize,journal]{IEEEtran}
\usepackage{amsmath,amsfonts}
\usepackage{algorithmic}
\usepackage{algorithm}
\renewcommand{\algorithmiccomment}[1]{\hfill $\triangleright$ #1}
\usepackage{color}
\usepackage{array}
\usepackage[caption=false,font=normalsize,labelfont=sf,textfont=sf]{subfig}
\usepackage{textcomp}
\usepackage{stfloats}
\usepackage{url}
\usepackage{verbatim}
\usepackage{graphicx}
\usepackage{cite}
\usepackage{hyperref}
\usepackage{booktabs}
\usepackage{multirow}
\usepackage{amsmath}
\usepackage{amssymb}
\usepackage{mathtools}
\usepackage{amsthm}
\usepackage{makecell}

\theoremstyle{definition}
\newtheorem{theorem}{definition}[section]

\theoremstyle{definition}
\newtheorem{definition}[theorem]{Definition}

\theoremstyle{remark}

\newcommand{\eg}{\textit{e.g., }}%
\newcommand{\ie}{\textit{i.e., }}%
\newcommand{\etal}{\textit{et al. }}
\newcommand{\etc}{\textit{etc}}
\newcommand{\method}[1]{CEM}

\begin{document}

\title{Interpreting Low-level Vision Models \\ with Causal Effect Maps}
\author{Jinfan Hu, Jinjin Gu, Shiyao Yu, Fanghua Yu, Zheyuan Li, Zhiyuan You, Chaochao Lu, Chao Dong

\thanks{This work was supported in part by the National Natural Science Foundation of China (Grant No. 62276251), the Joint Lab of CAS-HK, and the Shanghai Artificial Intelligence Laboratory. (Corresponding author: Chao Dong)}

\thanks{Jinfan Hu is with the Shenzhen Institutes of Advanced Technology, Chinese
Academy of Sciences, Shenzhen 518055, China, and also with the University of Chinese Academy of Sciences, Beijing 100049, China.  (e-mail: jf.hu1@siat.ac.cn)}

\thanks{Jinjin Gu and Chaochao Lu are with the Shanghai Artificial Intelligence Laboratory, Shanghai 200232, China. (e-mail: hellojasongt@gmail.com, luchaochao@pjlab.org.cn)}

\thanks{Shiyao Yu is with the Southern University of Science and Technology, Shenzhen 518055, China, and also with the Shenzhen Institutes of Advanced Technology, Chinese
Academy of Sciences, Shenzhen 518055, China. (e-mail: sy.yu1@siat.ac.cn)}

\thanks{Fanghua Yu and Zheyuan Li are with the Shenzhen Institutes of Advanced Technology, Chinese
Academy of Sciences, Shenzhen 518055, China. (e-mail: fanghuayu96@gmail.com, zy.li3@siat.ac.cn)}

\thanks{Zhiyuan You is with The Chinese University of Hong Kong, Hong Kong 999077, China, and also with the Shenzhen Institutes of Advanced Technology, Chinese
Academy of Sciences, Shenzhen 518055, China. (e-mail: zhiyuanyou@foxmail.com)}

\thanks{Chao Dong is with the Shenzhen Institutes of Advanced Technology, Chinese Academy of Sciences, Shenzhen 518055, China, and also with the Shanghai Artificial Intelligence Laboratory, Shanghai 200232, China. (e-mail: chao.dong@siat.ac.cn)}

}

\markboth{Journal of \LaTeX\ Class Files,~Vol.~14, No.~8, August~2021}%
{Shell \MakeLowercase{\textit{et al.}}: A Sample Article Using IEEEtran.cls for IEEE Journals}


\maketitle

\begin{abstract}
Deep neural networks have significantly improved the performance of low-level vision tasks but also increased the difficulty of interpretability.
A deep understanding of deep models is beneficial for both network design and practical reliability. 
To take up this challenge, we introduce causality theory to interpret low-level vision models and propose a model-/task-agnostic method called Causal Effect Map (CEM). 
With CEM, we can visualize and quantify the input-output relationships on either positive or negative effects. 
After analyzing various low-level vision tasks with CEM, we have reached several interesting insights, such as: (1) Using more information of input images (\textit{e.g.,} larger receptive field) does NOT always yield positive outcomes. 
(2) Attempting to incorporate mechanisms with a global receptive field (\textit{e.g.,} channel attention) into image denoising may prove futile. 
(3) Integrating multiple tasks to train a general model could encourage the network to prioritize local information over global context. 
Based on the causal effect theory, the proposed diagnostic tool can refresh our common knowledge and bring a deeper understanding of low-level vision models. Codes are available at \href{https://github.com/J-FHu/CEM}{https://github.com/J-FHu/CEM}.
\end{abstract}


\section{Introduction}
\label{intro}

\IEEEPARstart{T}{he} development of deep neural networks has significantly improved the performance of Low-level Vision (LV) tasks including image super-resolution, denoising, deraining, \etc.
Deep-learning-based methods claim that they integrate diverse finely crafted modules to realize various functions. 
However, we cannot actually determine whether they are functioning properly as we expect. 
Furthermore, general models \cite{SWinIR,MPRNet,HAT,XRestormer} are nowadays a major focus of this field, as they usually come with one unified architecture for different tasks.
Understanding whether networks behave differently when confronted with different tasks is beneficial.
To meet this challenge, network interpretability in low-level vision is receiving increasing attention~\cite{gu2023networks,gu2021interpreting,liu2021discovering,xie2021finding}.
Nevertheless, the interpretability of general models has not yet been discussed, since a general model requires a general interpretability method, which is not a property of previous methods.

More importantly, previous methods only confine their studies to correlations but neglect comprehensive investigations into causality.
It is widely acknowledged that ``correlation does NOT imply causation'', \ie a causal relationship between two events or variables cannot be legitimately deduced only from their correlation \cite{pearl2018book}.
Discovering the real cause of the network's output may be the centerpiece, rather than observing correlations \cite{gu2021interpreting,xie2021finding}.
Furthermore, the nuances of positive and negative causal relationships among variables remain unexplored. 
To sum up, the desired interpretability method could provide causal analysis and generalize well to diverse low-level models and tasks. 

\begin{figure}
    \centering
    \includegraphics[width=\linewidth]{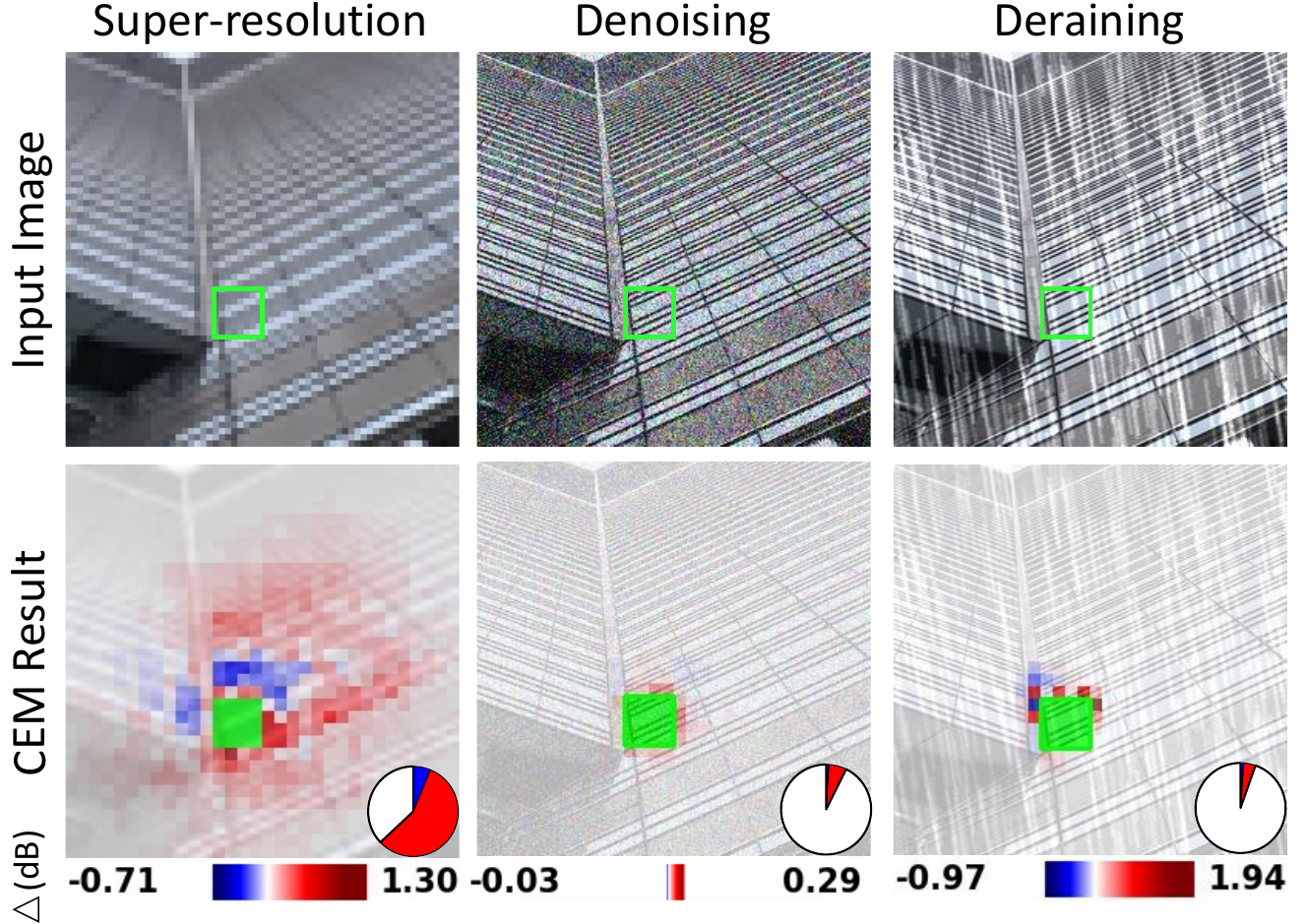}
    \caption{CEMs of SwinIR for the image super-resolution, denoising, and draining tasks. The patches with positive or negative causal effects and the ROI are indicated in red, blue, and green respectively. The pie chart records the percentage of patches with different causal effects, and the colorbar expresses the effect range.}
    \label{fig:teaser}
\end{figure}

To fill the gap, we propose a general interpretability method--the Causal Effect Map (CEM), serving as the first incorporation of causality theory into low-level vision interpretability. 
CEM aims to find the relationship between the patches in the input image and the Region Of Interest (ROI) in the output image.
Causal relationships can be revealed through visualization maps, while positive and negative causal effects can also be quantified.
It can be applied without any prior knowledge about the models, \ie we treat LV networks as black boxes, with any details of network architectures remaining blind to us.
Specifically, we conduct LV intervention on the patches of the input image appropriately (The definition and principles are presented in Sec. \ref{ss:LV intervention}).
The subsequent changes in ROI restoration performance can be recorded, which can be used to analyze the causality.

Rather than discerning the correlation between input and output images, we prefer to identify specific regions in the input image that positively or negatively affect the ROI's reconstruction.
As in Fig. \ref{fig:teaser}, CEM highlights those input patches that have causal effects on the ROI (green block).
Furthermore, the positivity and negativity of the effects are also illustrated in red and blue, respectively. 
With the help of CEM, we can analyze the network by calculating the percentage of positive and negative regions and the range of causal effects.
More importantly, CEM is a general framework that can help us perform cross-task interpreting, which is not possible with previous methods.
With the information brought by CEM, we have obtained some important insights.
\begin{itemize}
    \item The capacity to leverage an increased number of patches within input images does not invariably ensure improvement. 
    Involving a wider range of information also increases the likelihood of misguidance. 
    \item The network tends to only concentrate primarily on local areas when addressing the task of image denoising. 
    Those well-designed mechanisms with a global receptive field do not work as we expect.
    \item When incorporating multiple LV tasks to train a general model, the network slacks off and only focuses on local information, even though the model architecture is capable of collecting global knowledge.
\end{itemize}

Overall, we propose a model-/task-agnostic tool, CEM, that interprets LV models in a unified manner based on LV interventions. 
Moreover, we not only provide a comprehensive definition of LV interventions but also delineate their foundational principles.  
Building on this, we successfully introduce causality into LV interpretability and prove its superiority to traditional correlation-based approaches.

\section{Related Work}
\label{related}
\textbf{Low-level Vision:}
The LV tasks focus on computer vision tasks where images serve as input, processing, and output units.
The primary objective is to reconstruct or enhance low-quality images that suffer from diverse forms of degradation \cite{SRCNN, DnCNN, PReNet, 10843251, Xia_Hang_Tian_Yang_Liao_Zhou_2022, Xia_2023_ICCV}. 
Representative LV tasks, categorized by the specific type of degradation, encompass super-resolution (SR), denoising (DN), deraining (DR), and more.

With the recent development of deep learning, various LV models have emerged.
The performance and efficiency are enhanced by employing various techniques. 
One notable pipeline is the use of residual connection structures, which have been successfully implemented in several models such as CARN, EDSR, SRDenseNet, and SRResNet \cite{CARN, EDSR, SRDenseNet, SRResNet}. 
These structures help mitigate the vanishing gradient problem and facilitate the training of deeper networks, leading to improved accuracy and stability.
Another significant advancement in network design is the integration of attention mechanisms.
Models utilize these mechanisms to selectively focus on the most relevant features enhancing the network's ability to capture important information and improve overall performance \cite{RCAN, RIDNet, PAN}.
More recently, Transformers have opened up new possibilities for network architectures, enabling them to tackle a wider range of tasks with greater efficiency. 
Transformer-based models such as SCUNet \cite{SCUNet} and HAT \cite{HAT} leverage the power of self-attention, allowing for better handling of long-range dependencies. 

In addition to these advancements, it is important to recognize that low-vision (LV) tasks are often intertwined, and low-quality images may suffer from various degradations. 
To address this complexity, researchers have proposed several general LV models capable of solving multiple tasks within a single network backbone. 
Notable examples include SWinIR \cite{SWinIR}, MPRNet \cite{MPRNet}, and X-Restormer \cite{XRestormer}, which are designed to handle multitasking problems effectively, providing a comprehensive solution for various image restoration and enhancement challenges.

\textbf{Interpretability in Computer Vision:}
Interpretability has become an increasingly important topic in high-level vision research, with significant contributions from various studies \cite{sun2024explain,yosinski2015understanding,fong2017interpretable,zeiler2014visualizing,fong2019understanding,zhou2016learning}.
Zeiler \etal \cite{zeiler2014visualizing} provided an intuitive method to visualize the activity of convolutional networks, enhancing the understanding of their inner workings.
Zhou \etal \cite{zhou2016learning} introduced the concept of Class Activation Mapping (CAM), which enables the identification of discriminative regions in an image for a specific class. 
Sundararajan \etal \cite{sundararajan2017axiomatic} introduced Integrated Gradients, a method for attributing the predictions of deep networks to their input features while avoiding gradient saturation.
Fong \etal \cite{fong2017interpretable,fong2019understanding} found the most sensitive locations to models by perturbing different regions of the image and quantitatively measured the impact of these regions on the model results (classification probability).

The research on LV interpretability is still in its early stages, with only a few recent works focusing on this area.
The Local Attribution Map (LAM), proposed by Gu \etal \cite{gu2021interpreting}, illustrates the most relevant input regions to the output local region by integrating gradients.
Similarly, Xie \etal \cite{xie2021finding} identify filters that demonstrate high sensitivity to the network's output through the integrated gradient method.
Liu \etal \cite{liu2021discovering} conclude in their study that SR networks do not primarily focus on image content but exhibit a significant ability to distinguish between different types of degradation.
Magid \etal \cite{Magid_2022_CVPR} leverage a texture classifier to identify the source of SR errors both globally and locally.

\textbf{Causal Inference:}
Causal inference stands as an emerging discipline that aims to unveil and quantify the underlying causal relationships between variables \cite{pearl2022external,pearl2018book}. 
This theory has found widespread applications in various domains \cite{lu2018deconfounding,lu2021invariant,feder2022causal,qi2020two,yue2020interventional}.
AlterRep \cite{ravfogel2021counterfactual} generates counterfactual representations by changing the way linguistic features are encoded while preserving all other aspects of the original representations. 
By measuring changes in the predictive behavior of model words when these counterfactual representations replace the original representations, conclusions can be drawn about the causal effects of linguistic features on model behavior.
For the high-level vision tasks, counterfactual reasoning \cite{besserve2018counterfactuals}, and intervention-based analysis \cite{Wang_2020_CVPR_Workshops,schwab2019cxplain} hold the potential to illuminate the causal structures and dependencies within visual data.
Liu \etal introduce a novel image captioning framework called CIIC that addresses visual and linguistic confounders \cite{Liu_2022_CVPR}.
It employs an interventional object detector and an interventional Transformer decoder, significantly enhancing captioning performance.

Despite its successful application in those domains, the application of causality in LV tasks remains an unexplored frontier.
As the field of causal inference continues to evolve, its potential applications in LV tasks are expected to grow. 
Future research could explore the development of new causal inference methods specifically tailored for LV problems, as well as the adaptation of existing techniques to better suit the unique characteristics of LV input.

\begin{figure}[tpb]
    \centering
    \includegraphics[width=0.95\linewidth]{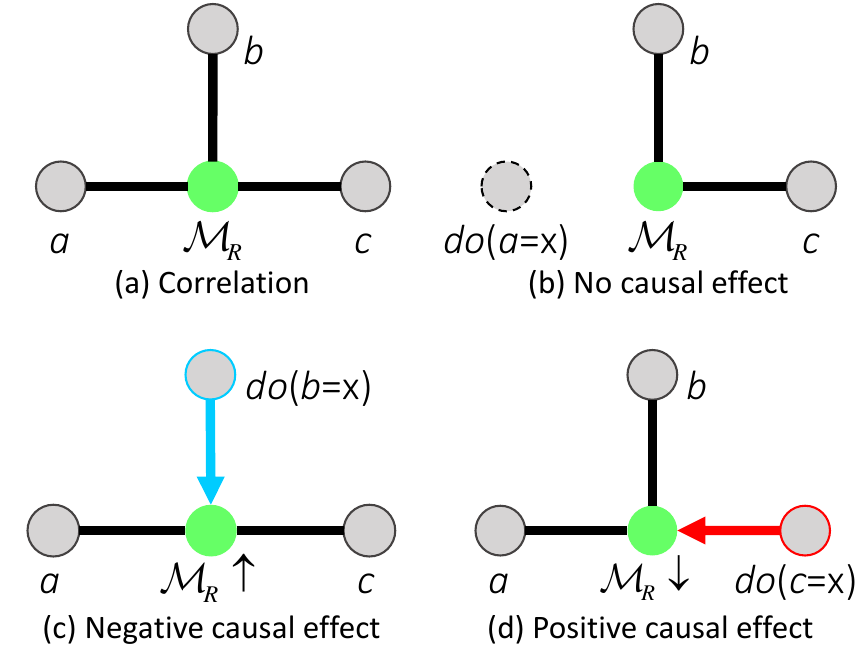}
    \caption{
    (a) Correlations between \textit{a}, \textit{b}, \textit{c} and \textit{R}, $\mathcal{M}_\textit{R}$ denotes some kind of measurement for \textit{R}.
    Intervening in \textit{a} ($\textit{do}(a=x)$), \textit{b} ($\textit{do}(b=x)$), and \textit{c} ($\textit{do}(c=x)$) reveals that (b) variable \textit{a} has no causal relationship with \textit{R}, (c) variable \textit{b} has a negative causal effect, while (d) variable \textit{c} has a positive causal effect.
    The black undirected edge represents two variables that correlate. The blue-directed and red-directed edges show the negative and positive causal relationship, respectively.
    }
    \label{fig:corr1}
\end{figure}

\section{Motivation: From Correlation to Causation}
It is well known that ``correlation does not imply causation.'' 
\ie the mere observation of a correlation between two events or variables does not justify the existence of a causal relationship \cite{verhulst2012correlation,pearl2022external,pearl2010causal}. 
Despite this well-established principle, previous LV interpretability methods predominantly anchored on correlation analysis \cite{gu2021interpreting,xie2021finding}, overlooking the critical distinction between correlation and causation.

Consider, for example, the sales of T-shirts and ice cream during the summer. 
One might observe a simultaneous increase in the sales of both items, indicating a clear positive correlation between them.
Imagine a scenario where we have the ability to intervene to equalize T-shirt sales between summer and winter, \ie \textit{do}(summer sales of T-shirts = winter sales of T-shirts). 
Here, $\textit{do}(p=x)$ means fixing the variable $p$ to the value $x$ while maintaining the rest of the model unaltered \cite{pearl2022external}. 
This operator is instrumental in isolating the specific effects of $p$ on the model's output, thereby facilitating a focused analysis of causal influence within the system.
By implementing this intervention, it becomes apparent that ice cream sales still increase during the summer even with T-shirt sales remaining steady, indicating the absence of a causal relationship between them.
Generally speaking, correlation overshadows causation.
The above example also illustrates that intervention-based approaches can delve into deeper relationships.

Local Attribution Map (LAM) is the first interpretation method in image SR \cite{gu2021interpreting}.
It adeptly highlights regions within the input image that strongly correlate with the ROI in the output of SR networks.
An examination can be carried out to determine if mechanisms that augment receptive fields improve pixel utilization in the prediction process.
Nonetheless, the direct impact of these highlighted regions on network prediction, along with the nature of their contribution - whether positive or negative - to the final outcome, remains unclear.
Fig. \ref{fig:corr1} illustrates the differences between correlation and causation.
As shown in Fig. \ref{fig:corr1}(a), variables \textit{a}, \textit{b}, and \textit{c} are all correlated with \textit{R}.
Intervening on variables \textit{a}, \textit{b}, and \textit{c}, as shown in Fig. \ref{fig:corr1}(b)-(d), will manifest distinct outcomes according to their causal relationships with \textit{R}.
\begin{figure}[tpb]
    \centering
    \includegraphics[width=\linewidth]{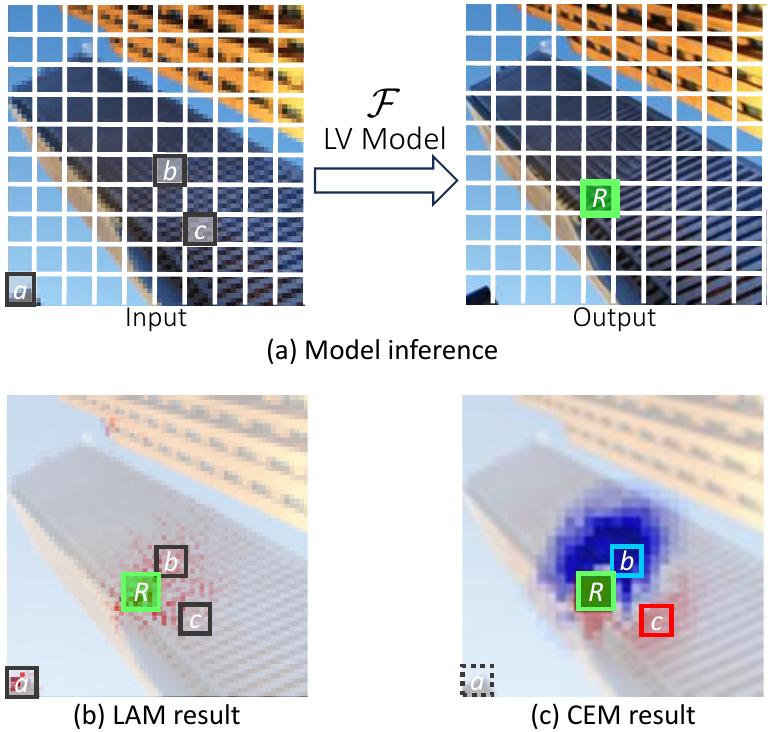}
    \caption{
    (a) Examining the relationship between input patches \textit{a}, \textit{b}, \textit{c} and the reconstruction of ROI \textit{R}.
    (b) The LAM of RNAN, red dots represent the correlation. (c) The CEM of RNAN, patches with negative and positive effect are highlighted in blue and red.
    }
    \label{fig:corr2}
\end{figure}

In the context of interpreting image super-resolution models, Fig \ref{fig:corr2} provides a specific example.
Upon examining the intricate relationship between input patches \textit{a}, \textit{b}, \textit{c}, and the reconstruction of \textit{R} by RNAN \cite{RNAN}, correlations among \textit{a}, \textit{b}, \textit{c}, and \textit{R} are observed in the LAM result shown in Fig. \ref{fig:corr2}(b).
This initial observation often leads us to assume that these three patches collectively contribute to \textit{R}. 
However, with closer inspection and through individual interventions on patches \textit{a}, \textit{b}, and \textit{c}, the underlying causal effects on \textit{R} begin to be revealed.
Shifting attention to our CEM in Fig. \ref{fig:corr2}(c), 
it reveals the positive or negative effect of corresponding patches on the reconstruction of \textit{R}.
To sum up, patches \textit{a}, \textit{b}, and \textit{c} are all correlated with \textit{R}, but they are not causally identical.

\section{Methods}
\label{methods}

As mentioned in the introduction, we favor an interpretability method capable of not only offering causality analysis but also exhibiting strong generalizability across diverse LV tasks.
Thus, we propose a general interpretability method for various LV models, the Causal Effect Map (CEM).
We first introduce the framework of general intervention in LV in Sec. \ref{ss:framework}, followed by the definition and principles of LV interventions in Sec. \ref{ss:LV intervention} and \ref{ss:pricinple}.
Based on the LV interventions, we propose CEM to reveal the causality of LV models in Sec. \ref{ss:CEM}.
Finally, we present a coarse-to-fine pipeline for accelerating CEM calculations in Sec. \ref{ss:accelrating}.

\subsection{General Intervention Framework}
\label{ss:framework}

The current challenge is to develop a framework that meets the requirements of interpreting diverse networks across various tasks.
Previous approaches \cite{gu2021interpreting,gu2023networks} rely on network structure or task characteristics, which has led to an inability to analyze across tasks.
Thus, our goal is to develop an interpreting approach that is both model-agnostic and task-agnostic.
In other words, we must treat all models as black boxes.
Consequently, our focus shifts to the intrinsic commonality across models: the input-output relationship.

We also need to clarify the interpretability principles in LV, which are proposed in LAM 
\cite{gu2021interpreting}, \ie focusing on the restoration of local regions with texture.
Specifically, the output images of LV networks are spatially corresponding to the whole input image.
The entire input image can be therefore considered a determinant cause of the network's generation of the corresponding output image.
Global explanations do not yield discerning information.
Additionally, the restoration of flat regions without texture is relatively simple for LV models, focusing on these regions cannot distinguish their performance and behaviors.
The goal of our method is to find the causal relationship between the input image patches and the reconstruction of challenging ROIs in the output image.

Inspired by the notion of intervention in causality \cite{pearl2018book}, we find that the method of interpreting the network through interventions (or perturbations) dovetails seamlessly with our objective.
The intervention serves as a direct and effective method for assessing the existence of a direct causal relationship between variables \cite{pearl2022external,pearl2018book}.
When examining the causal relationship between a variable $p_i$ from the input set $I$ and its corresponding output $\mathcal{F}(I)$ through the function $\mathcal{F}(\cdot)$, we can disrupt the observed correlation of the original input and output by intervention, denoted as $\textit{do}(p_i=x)$. 
Then, we can analyze the subsequent change as follows:
\begin{equation}
\label{eq:CE}
\begin{aligned}
    \phi_\mathcal{F}[p_i] &=\mathcal{M}(\mathcal{F}(I)) - \mathcal{M}(\mathcal{F}(I|\textit{do}(p_i=x)))\\
    &=\mathcal{M}(\mathcal{F}(I)) - \mathcal{M}(\mathcal{F}(I')),
\end{aligned}
\end{equation}
where $\phi_\mathcal{F}[p_i]$ can be regarded as the causal effect of $p_i$ through the function $\mathcal{F}$.
$I'$ denotes the intervened inputs.
$\mathcal{M}$ can be any function that measures the effect of an outcome. 
Using Eq. \ref{eq:CE} to traverse the entire input set $I$, we can obtain the causal effect map of this set, \ie $\phi_\mathcal{F}(I)=\{\phi_\mathcal{F}[p_i]\}_{i=1}^N$. $N$ denotes the total number of patches in the input image $I$.

In the context of LV, $p_i$ signifies a patch in the input image, $I$ represents the original input image, $\mathcal{F}$ denotes the network, and $\mathcal{M}_R$ ( \ie $\mathcal{M}$ in Eq. \ref{eq:CE}) signifies the assessment of image quality for a selected ROI $R$.
Note that, intervening with patches within the ROI
would disrupt their original content, making the analysis meaningless. 
Thus our focus is on patches outside the ROI.
At the same time, since the patches within the ROI are the direct contributors to the ROI restoration, we define the causal effect of those patches as $+\infty$.
To sum up, the pipeline can be described as Alg. \ref{alg1}.
\begin{algorithm}[tb]
    \footnotesize
   \caption{Intervention Framework for LV models}
   \label{alg1}
    \begin{algorithmic}[1]
   \STATE \textbf{Input:}  $I$ (input image) ,  $x$ (intervention patch), $\mathcal{F}$ (LV model), $\mathcal{M}_R$ (performance measurement).
   \STATE \textbf{Output:} $\phi_\mathcal{F}(I)=\{\phi_\mathcal{F}[p_i]\}_{i=1}^N$ \algorithmiccomment{Causal effect map of $I$}
   \STATE $M_o = \mathcal{M}_R(\mathcal{F}(I))$ \algorithmiccomment{Original performance}
   \FOR{$p_i$ outside $R$}
   \STATE  $I' = I|\textit{do}(p_i=x)$ \algorithmiccomment{Intervening $I$ with $x$}
   \STATE  $M' = \mathcal{M}_R(\mathcal{F}(I'))$ \algorithmiccomment{Post-intervention performance}
   \STATE  $\phi_\mathcal{F}[p_i] = M_0 - M'$ \algorithmiccomment{ Causal effect of variable $p_i$}
   \ENDFOR
   \FOR{$p_i$ inside $R$}
   \STATE $\phi_\mathcal{F}[p_i] = +\infty$ \algorithmiccomment{ Causal effect of variable $p_i$}
   \ENDFOR
\end{algorithmic}
\end{algorithm}
\subsection{Defining LV Interventions}
\label{ss:LV intervention}

Comparing the pre-intervention input $I$ with the post-intervention input $I'$, the changes observed in the output could reveal its causal relationship with the examined variable $p_i$.
The upcoming key question we need to address is how to define LV interventions and what their purpose is.

LV encompasses various tasks like super-resolution, denoising, deraining, \etc.
Degradation represents the inherent characteristic of various tasks and serves as a key differentiator among them. 
The premise of the LV network's proper performance is that the input image falls in a degraded distribution that the network has learned before.
Therefore, we believe that when LV interventions are performed, the degradation information in the post-intervention image needs to align with the original image.
Our focus is on how the model utilizes the image content behind these degradations.

\begin{definition}[\textbf{LV Intervention}]
\label{def:intervention}
Given an input image $I$ that contains a patch $p_i$ and an alternative patch $x$, the intervened image $I'$ is generated by applying the $do$ operator to $p_i$ with $x$, denoted as $I' =I|\textit{do}(p_i=x) $.
An intervention is considered valid in LV tasks if $I$ and $I'$ still lie in the same degradation distribution.
\end{definition}

Referring to Def. \ref{def:intervention}, we can clarify that the intervention in LV solely deals with the content of the image. 
This specifies the goal of the LV intervention, which is to change the image content while ensuring that the image degradation type remains unchanged after the intervention.

\subsection{Characterizing LV Interventions} \label{ss:pricinple}
\begin{figure}[t]

    \centering
    \includegraphics[width=\linewidth]{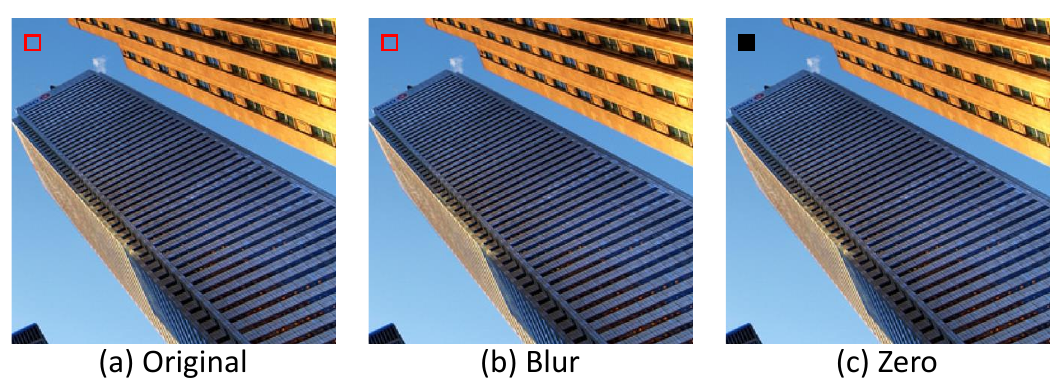}

    \caption{The patch in the original image (a) does not exhibit a significant difference compared to the blur intervention in (b). Assigning zero values in (c) causes the patch to deviate from its neighbors.}
    \label{fig:intervention}

\end{figure}

The choice of a suitable intervention method is a prerequisite for conducting all subsequent analyses accurately.
Here we propose two principles of intervention in LV:

\textbf{(i) An intervention should be meaningful. }
It should effectuate substantive alterations in the content of input patches.
The intervention needs to represent a difference from the original variable at a certain level.
Within the domain of computer vision, prevalent intervention techniques applied to images include blurring, noise addition, and the assignment of zero or other constant values to pixels \cite{fong2017interpretable,ancona2017towards}.
They use the above interventions to produce some sort of difference between variables before and after the intervention.
What stands out through blurring intervention is the information between the image and its blurred counterpart. 
It actually simulates the absence of high-frequency information.
However, the appropriateness of blurring as a \textit{do} operator is not universally applicable across all LV tasks.
Specifically, the blurring operation is rendered ineffectively in scenarios where the patch content is flat and simple, as illustrated in Fig. \ref{fig:intervention}.
The difference between Fig. \ref{fig:intervention}(a) and (b) cannot be discovered.

\textbf{(ii) An intervention should be harmless. }
It should ensure that the image patch maintains its original distribution.
For example, directly erasing all content within an image patch (\ie set $\textit{do}(p_i=0)$) is a common intervention in high-level vision tasks. 
For the high-level vision tasks of extracting abstract knowledge from images (\eg getting class probabilities from images), they are insensitive to the local details of the image but focus on the overall semantics. 
Removing a fraction of the image via $\textit{do}(p_i=0)$ is acceptable for high-level vision.

However, LV tasks are particularly sensitive to changes in pixel intensity.
The abrupt introduction of one simple all-black patch - an anomaly starkly contrasted against its surroundings - can result in a significant departure from the natural image distribution, as depicted in Fig. \ref{fig:intervention}(c).
Such interventions can disrupt the normal behavior of LV networks, leading to significant changes that may render subsequent analyses unreliable.

An intuitive method to assess the significance of a variable is to analyze the effect of altering the variable's contribution, either by increasing or decreasing it from its original state, on the model's output.
For instance, a coffee shop can quantitatively evaluate the impact of the coffee price on sales by observing the changes in sales volume following an increase or decrease in the price, with other factors held constant.
In the LV domain, it is challenging to clearly identify the contributions of specific patches and enhance them in a practical way.
The ablation of patches, on the other hand, may be easier to achieve and more reliable.
Therefore, our intervention strategy is to address the scenario where the variable is in an absence state, presenting an average effect in which the variable has no merit or fault.
Building on this understanding, Eq. \ref{eq:CE} can be regarded as calculating the causal effect difference between the original state \(\mathcal{M}_R\mathcal{F}(I)\) and the absence state \(\mathcal{M}_R(\mathcal{F}(I \mid \text{do}(p_i = x)))\).

\begin{figure*}[htp]
    \centering
    \includegraphics[width=\linewidth]{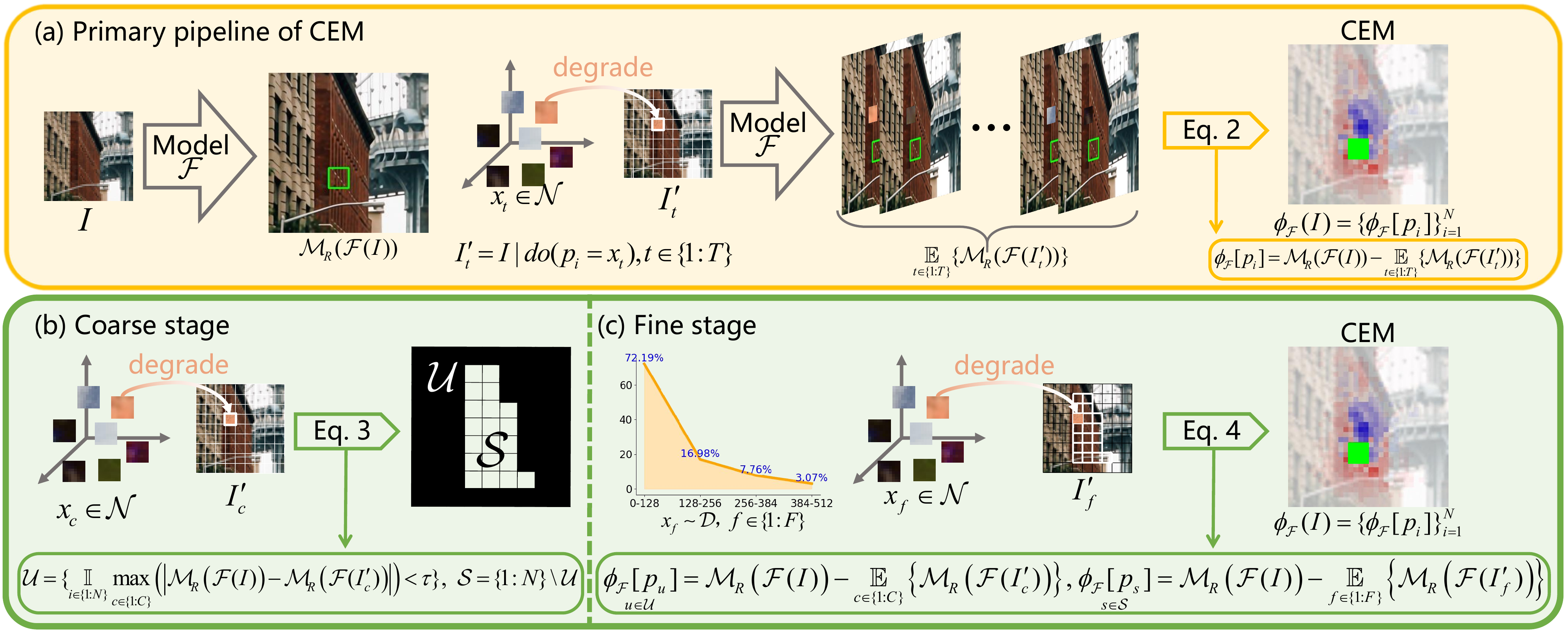}
    \caption{(a) Intervening all the patches for $T$ times to get a CEM. (b) Dividing the patches into two sets $\mathcal{U}$ and $\mathcal{S}$ in the coarse stage. (c) Refining the causal effect of set $\mathcal{S}$ in the fine stage. The patches are sampled according to the probability density $\mathcal{D}$, which is further discussed in Sec. \ref{A-accel}.}
    \label{fig:method}
\end{figure*}
\subsection{Causal Effect Map for LV Models}
\label{ss:CEM}

To satisfy the above principles, we adopt an approach to averaging over multiple interventions. 
The single-specific intervention like blurring and zero-value assignment has its own limitations in that they do not lead to changes in image content or disrupt the original distribution.
We cannot explicitly determine how a patch not contributing specially to the model should look like.
So we chose to randomly crop patches from the natural image library for multiple interventions, taking the impact from these interventions as an average impact from the natural image library.
Specifically, we choose the widely spread natural image dataset DIV2K \cite{Timofte_2018_CVPR_Workshops} as the intervention image library $\mathcal{N}$ (More discussions about the selection of natural image library are in Sec. \ref{A-dataset}).
By performing a set of calculations using different intervention patches $x_t$ from $\mathcal{N}$, a mathematical expectation is obtained, which is known as the Average Treatment Effect (ATE) \cite{angrist1995identification,yao2021survey,shalit2017estimating,vanderweele2013causal}.
It is a widely used metric for evaluating the impact of a treatment or intervention on subjects in an experiment. 

ATE is particularly important in studies focused on causal inference because it quantifies the causal effect by representing the difference between the average outcomes of the treated group and the untreated group. 
It provides the average causal effect of the treatment across the entire population, beyond the individual level. 
This helps to eliminate individual differences introduced by interventions, giving a clearer picture of the treatment's overall impact.
When investigating the effect of a drug, dividing the population into treated and untreated groups to analyze the results is a notable example of ATE.

In our scenario, the treated group is \(\mathcal{F}(I)\), and the untreated group, which represents the absence state of patch \(p_i\), is \(\mathcal{F}(I|do(p_i=x_t))\).
The whole ATE calculating process in LV tasks can be described as:
\begin{equation}
\begin{aligned}
    \phi_\mathcal{F}[p_i] &=\operatorname{ATE}_\mathcal{F}[p_i] \\
    &= {\mathbb{E}}\{\mathcal{M}_R({\mathcal{F}}(I))\} - \mkern-10mu\underset{t\in \{1:T\}}{\mathbb{E}}\{\mathcal{M}_R(\mathcal{F}(I|do(p_i=x_t)))\} \\
    &= \mathcal{M}_R({\mathcal{F}}(I)) - \mkern-10mu\underset{t\in \{1:T\}}{\mathbb{E}}\{\mathcal{M}_R(\mathcal{F}(I'_t))\},
\end{aligned}
\label{eq:500}
\end{equation}
where $p_i$ denotes a patch in the input image, $T$ indicates the number of interventions, $\mathcal{M}_R$ represents the function measuring image quality (PSNR in this work) for the chosen ROI, and $\mathcal{F}$ denotes the LV model. 
$do(p_i=x_t)$ indicates the replacement of $p_i$ with $x_t$ that suffers the same degradation as $p_i$.
The whole process is shown in Fig. \ref{fig:method}(a).

Due to the inherent specificity of the LV tasks, it is challenging to isolate the variable $p_i$ from other variables completely.
When intervening on $p$, it inevitably influences the surrounding pixels. 
To mitigate this impact, we crop small patches of size $8 \times 8$ in an input image of size $256 \times 256$ in this study (More discussions about the patch size are in Sec. \ref{A-size}), \ie a single image is cropped into $32\times32 = 1024$ patches.
Substituting Eq. \ref{eq:500} into Alg. \ref{alg1}, we can then obtain the result $\phi_\mathcal{F}(I)$, \ie CEM of the model $\mathcal{F}$ with the image $I$.

\subsection{Accelrating CEM Calculation}

\label{ss:accelrating}
\begin{figure}
    \centering
    \includegraphics[width=\linewidth]{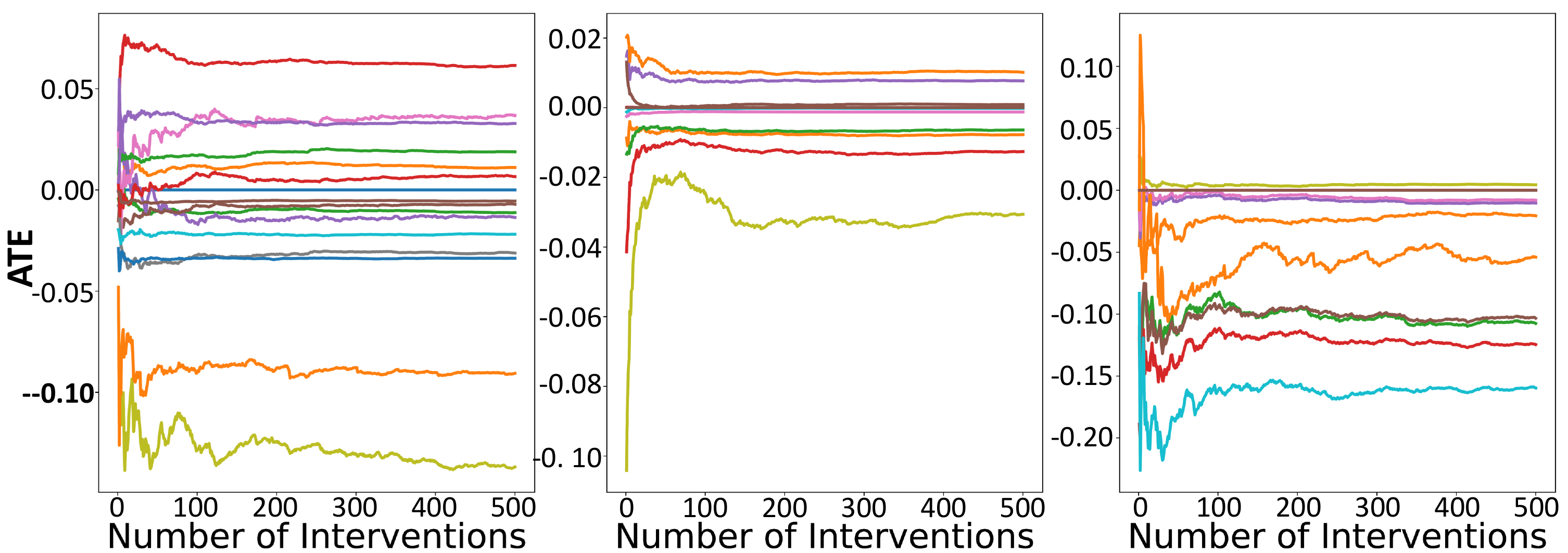}
    \caption{Examples showing that the ATE can converge as the number of interventions increases. Each line represents a SR model.}
    \label{fig:converge}
\end{figure}

Following the computations outlined in Fig. \ref{fig:method}(a), we can generate CEM $\phi_\mathcal{F}(I)$ of model $\mathcal{F}$ corresponding to image $I$.
However, calculating a stable causal effect requires multiple interventions ($T=500$ in our experiments). 
Fig. \ref{fig:converge} records the convergence curves of the causal effect for three different patches, where each line represents an SR model. 
From this, it can be seen that the calculation of the patch's causal effect necessitates a large number of model inferences.

An image of $256 \times 256$ can be divided into 1024 patches of $8\times 8$.
Consequently, the total number of inferences needed for one image is $32\times32\times500=512000$, which poses a significant computational burden.
Adopting this pipeline, even networks as small as SRCNN \cite{SRCNN} take more than an hour to produce a CEM when inferring on one NVIDIA A6000 GPU, while larger Transformer-based networks like HAT \cite{HAT} take approximately ten hours to complete inference on these intervention images.
This approach is highly inefficient.

To accelerate the process, we propose a streamlined approach to CEM calculation, adopting a coarse-to-fine strategy. 
Initially, we applied a small number ($C$) of interventions to each patch in the first stage.
Because patches with strong causal effects on ROIs are usually also very sensitive to interventions, $C$ interventions are enough to find them out.
If the maximum absolute effect difference from all $C$ ($C=3$ in our paper) perturbations is less than a tolerance $\tau$ ($\tau=0.01$ dB in our paper when $\mathcal{M}_R$ is the PSNR), we categorize the patch as a set $\mathcal{U}$ that is unrelated to the ROI.
Conversely, patches that demonstrate a notable response to the intervention are incorporated into the sensitive set $\mathcal{S}$, as depicted in Fig. \ref{fig:method}(b). 
This process can be represented by the following equations:
\begin{equation}
\label{eq:coarse}
\begin{aligned}
\mathcal{U}&=\{\underset{i\in\{1:N\}}{\mathbb{I}}(\mkern-6mu \max_{c\in\{1:C\}}\mkern-6mu\mid\mkern-6mu \mathcal{M}_R(\mathcal{F}(I))\!-\!\mathcal{M}_R(\mathcal{F}(I'_{c}))\mid\mkern-2mu <\mkern-6mu \tau)\}, \\
\mathcal{S} &= \{1:N\}\backslash \mathcal{U}, \\
\end{aligned}
\end{equation}
where $\mathbb{I}$ represents the index indicator, recording the index when the condition is met. 
$N$ represents the total number of patches in the input image, $I'_{c} = I|do(p_i=x_c)$.

In the second stage, interventions are exclusively performed on patches within the set $\mathcal{S}$ that exert a causal effect on $R$. 
$F$ times of intervention ($F=50$ in our setting) are executed to attain a more stable causal effect, as shown in Fig. \ref{fig:method}(c).
The process can be presented through the following equations:
\begin{equation}
\begin{aligned}
    \underset{u\in\mathcal{U}}{\phi_\mathcal{F}[p_u]}&=\mathcal{M}_R(\mathcal{F}(I)) \mkern-3mu- \mkern-12mu \underset{c \in \{ 1:C \}}{\mathbb{E}}\mkern-3mu\{\mathcal{M}_R(\mathcal{F}(I'_{c}))\}, \\
    \underset{s\in\mathcal{S}}{\phi_\mathcal{F}[p_s]}&=\mathcal{M}_R(\mathcal{F}(I)) \mkern-3mu- \mkern-12mu \underset{f \in \{ 1:F \}}{\mathbb{E}}\mkern-3mu\{\mathcal{M}_R(\mathcal{F}(I'_{f}))\}.
\end{aligned}
\label{eq:fine}
\end{equation}
%
Furthermore, we collected the gradient information from patches in the set $\mathcal{N}$ to establish the probability density $\mathcal{D}$.
Sampling of intervention patches in the fine stage follows the probability density $\mathcal{D}$ (More discussions about the intervention strategy are presented in Sec. \ref{A-accel}). 
In this way, we ensure that those samples still conform to the distribution of natural images.
After accounting for 280 CEMs from different SR networks, 
we find that the proposed coarse-to-fine method (Fig. \ref{fig:method}(b) and (c)) uses only 4.9\% of the initial inference times on average compared to the primary pipeline presented in Fig. \ref{fig:method}(a).
CEMs obtained by the two strategies have an average similarity score $S$ of 82.6\% ($S=100\%-\|\phi'_\mathcal{F}(I)-\phi_\mathcal{F}(I)\|_1/\|\phi_\mathcal{F}(I)\|_1$, $\|\cdot\|_1$ is the $\ell_1$ norm).

To implement CEM, we have undertaken the following steps overall:
Firstly, we opt to intervene on sufficiently small patches to prevent significant deviations from the global semantic information in natural images (The patch size of $x_t \in \mathcal{N}$ in Fig. \ref{fig:method}(a) is relatively small).
Secondly, since the original structure contains both image content and degradation, and low-level models typically operate based on the type of degradation, we confine changes to the image content domain while keeping the degradation fixed ($I'_t$ in Fig. \ref{fig:method}(a) is generated with the degraded patch $x_t$).
Thirdly, we utilize multiple interventions instead of a single intervention to achieve a relatively accurate and stable causal effect from the ATE (Eq. \ref{eq:500}), eliminating biases introduced by individual intervention.
Furthermore, we propose a coarse-to-fine method to accelerate the computation of CEM.
Unrelated set $\mathcal{U}$ and sensitive set $\mathcal{S}$ are obtained as shown in Fig. \ref{fig:method}(b).
The causal effect of the sensitive set $\mathcal{S}$ is refined as shown in Fig. \ref{fig:method}(c).

\section{Experiments}
\subsection{Experiment Settings}
\label{setting}

In order to comprehensively evaluate diverse LV models, we collect several networks with different representative mechanisms.
Models in the experiments encompass basic CNN and residual networks: SRCNN \cite{SRCNN}, EDSR \cite{EDSR}, CARN \cite{CARN}, SRDenseNet \cite{SRDenseNet}, SRResNet \cite{SRResNet}, DnCNN \cite{DnCNN}, FFDNet \cite{FFDNet}, SADNet \cite{SADNet}, PReNet \cite{PReNet}, and HINet \cite{HINet};
attention mechanisms: DRLN \cite{DRLN}, RNAN \cite{RNAN}, RCAN \cite{RCAN}, SAN \cite{SAN}, and MPRNet \cite{MPRNet};
Transformer-based networks:  SwinIR \cite{SWinIR}, SCUNet \cite{SCUNet}, HAT \cite{HAT}, and X-Restormer \cite{XRestormer}. 
%
For the task setup, we select three representative tasks \ie super-resolution (SR), denoising (DN), and deraining (DR).
The low-resolution images are downsampled 4$\times$ by bicubic interpolation; noisy images are with the Gaussian noise level $\sigma=50$; and we follow the medium rain setting of previous work \cite{gu2023networks} to synthesize rainy images.
For the model weights, we chose to utilize the carefully trained official weights provided by the original authors to ensure robust network performance.
Furthermore, when it comes to model retraining, we uniformly employ the widely used dataset DIV2K as the training dataset.
As for the test images, we adopt the test set from LAM containing 150 challenging cases for LV models for all experiments.

\subsection{Mechanism Diagnosis}
Generally speaking, it is commonly believed that utilizing more pixels leads to better performance. 
A major motivation for LV network design is to enhance the network's ability of pixel utilization.
It has also been shown in LAM that there is a significant positive correlation between pixel usage and SR network performance.
However, the question arises: \textit{\textbf{Does all the information utilized truly have a positive influence on the model's outputs?}}  
\begin{figure}[t]
    \centering

    \includegraphics[width=\linewidth]{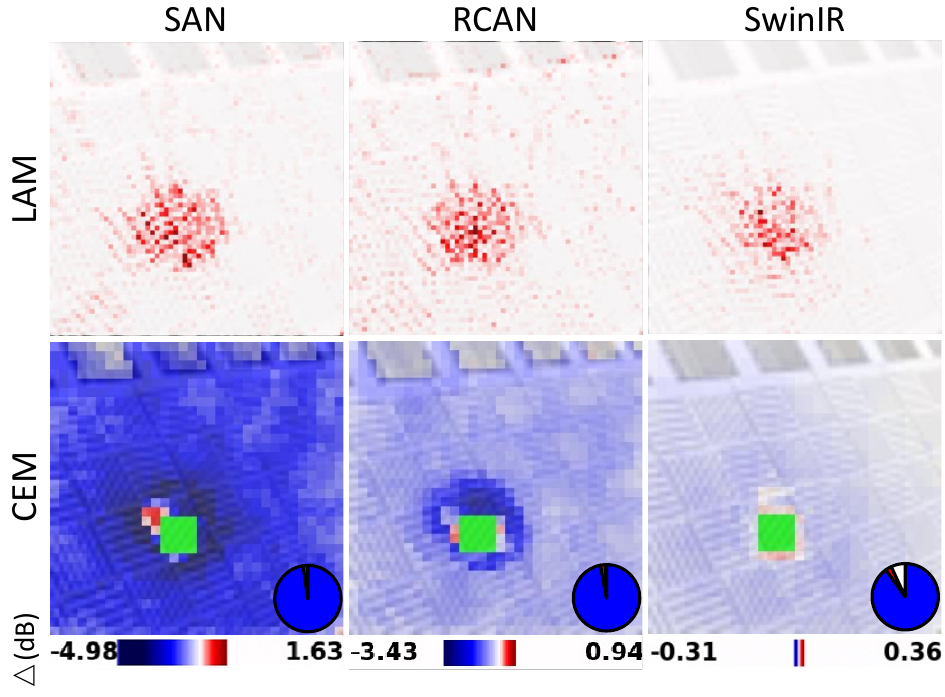}

    \caption{LAM and CEM results of SAN, RCAN, and SwinIR. Red dots in LAM indicate pixels correlate to the model output. Red and blue patches in CEM respectively represent those patches that have positive and negative causal effects on the output. The green area represents the ROI. The colorbar shows the causal effects range. The pie chart counts the percentage of positive and negative patches in the input image with red and blue.}
    \label{fig:LAMvsCEM}

\end{figure}

\begin{figure}[t]
    \centering
    \includegraphics[width=\linewidth]{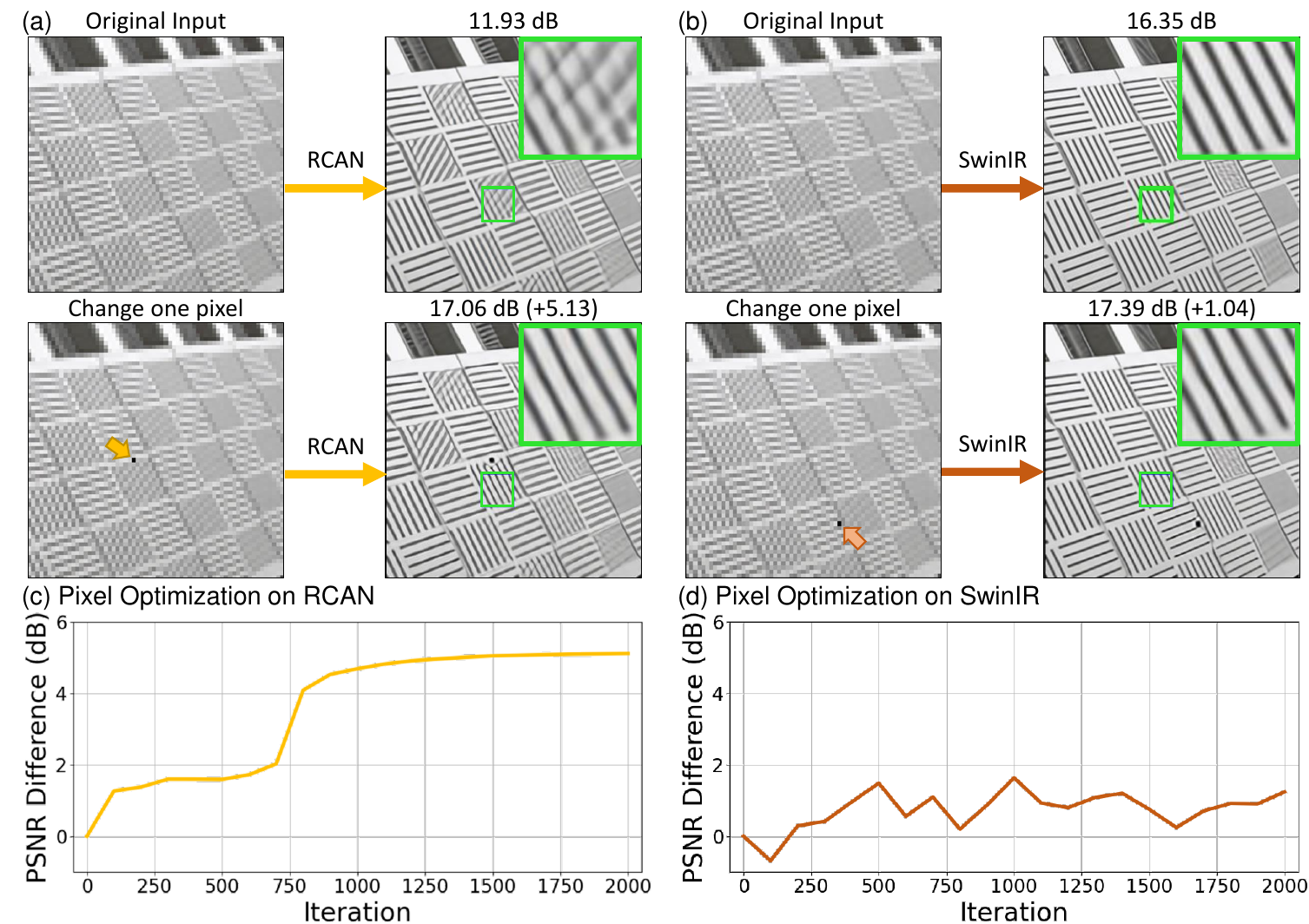}
    \caption{(a) Changing just one pixel in the most negative patch significantly alters the performance of RCAN, whereas (b) the performance change for SwinIR is relatively smaller. These pixels' RGB values are updated to (0, 0, 0) over 2000 iterations. They are located in their own most negative patches, respectively. PSNR Improvement During Pixel Optimization Iterations for (c) RCAN and (d) SwinIR.}
    \label{fig:onepixel}

\end{figure}

Fig. \ref{fig:LAMvsCEM} compares the LAM and CEM results of SAN, RCAN, and SwinIR, demonstrating the importance of causality.
From the LAM results, one might assume that all these networks effectively handle the image due to their extensive utilization of the input. 
However, the CEM results further unveil the underlying issues: a significant portion of the input patches have negative effects, misleading the outputs of different networks to varying degrees.
For SAN and RCAN, the most negatively contributing patch causes the PSNR to drop by 4.98 dB and 3.43 dB, respectively.
In contrast, for SwinIR, the input patches result in at most a 0.31 dB drop.
This demonstrates that SAN and RCAN are more vulnerable to being misled by the information in the input image compared to SwinIR.

We next design an experiment to explore whether modifying the most negatively contributing patch can improve the network’s performance and to what extent such improvement can be achieved.
In this experiment, a single pixel from the most negative patch is selected and set as a learnable parameter.
The loss function is defined as the L2 loss between the output and the ground truth (GT) ROI.
Using the Adam optimizer, this pixel is updated over 2000 iterations to enhance the reconstruction performance.
In the case of RCAN, modifying just one pixel in the most negative patch significantly increases the PSNR by 5.13 dB and corrects the erroneous texture, as shown in Fig. \ref{fig:onepixel}(a).
In contrast, for SwinIR, the original input pixel does not severely mislead the reconstruction process.
Updating the pixel does not result in a qualitative transformation of the reconstruction, and the performance improvement is smaller compared to RCAN, as illustrated in Fig. \ref{fig:onepixel}(b).
Moreover, as shown in Fig. \ref{fig:onepixel}(c) and (d), RCAN’s performance is more sensitive to pixel modifications and is more significantly improved by such updates compared to SwinIR.
This further demonstrates that RCAN is more susceptible to the influence of input information, with the original pixel contributing to greater negative effects on its reconstruction process.

\begin{figure}[t]
    \centering
    \includegraphics[width=\linewidth]{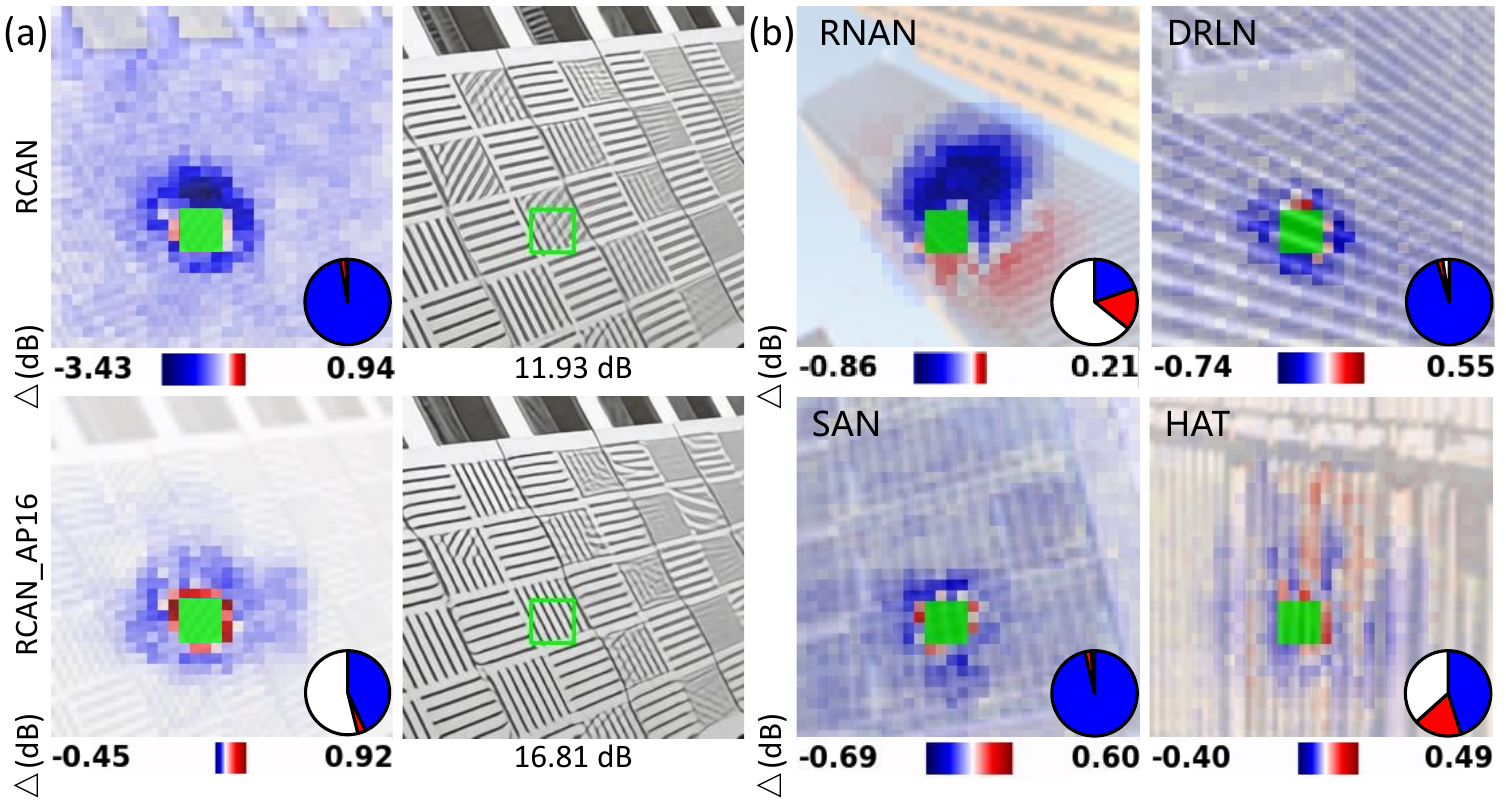}

    \caption{(a) Replacing global pooling with average pooling with a kernel size of 16 makes RCAN more stable. (b) More CEMs about SR networks involving many patches with negative causal effects. }
    \label{fig:rcan}

\end{figure}

Further analysis reveals that the global pooling layers in RCAN are the main cause of this problem.
Compressing feature maps of $H \times W \times C$ to vectors of $ 1 \times 1 \times C$ results in too much information loss, which is fatal for image restoration.
We replace global pooling with average pooling (kernel size 16) to obtain RCAN$\_$AP16 and observe that this issue is alleviated, as shown in Fig. \ref{fig:rcan}(a). 
The most negative causal effect caused by a single patch is diminished by a lot.
%
Fig. \ref{fig:rcan}(b) presents more CEMs of different SR networks being misled.
We cannot assume that the wide range of input pixels utilized by the networks will always necessarily have a positive impact.
As the input image utilization increases, the possibility of the network being tricked also increases.
Furthermore, we collect all the CEMs of SR networks tested with the test set and record the statistics in Fig. \ref{fig:summary}(a).
It is clear to observe that the range of input information utilized by SR networks is indeed expanding as the technology develops, but not all input pixels can bring positive contributions.
\textbf{\textit{Networks involve more pixels while also increasing the possibility of being misled.}}
Improving the network's ability to discriminate input information is probably the next major focus of studies.
\subsection{Differences Between LV Tasks}
In the previous subsection, we mainly focused on the SR model and explored the effect of input image patches on ROI.
Next, we will analyze DN and DR tasks and explore their similarities and differences with SR.
To eliminate the influence of model architecture and focus solely on the differences in tasks, we also employ the same structures to handle different tasks.
SwinIR is a general network for image restoration, the model weights of SR and DN are provided by the official.
In addition, we used DIV2K to generate synthetic datasets to retrain SRResNet, RCAN (for DN and DR), and SwinIR (for DR).
Generally, most of the LV networks are pursuing large receptive fields, expecting to include more information.
More importantly, we also wonder: \textit{\textbf{Do these networks truly perform as effectively as desired across all tasks?}}

\begin{figure}[t]
    \centering
    \includegraphics[width=\linewidth]{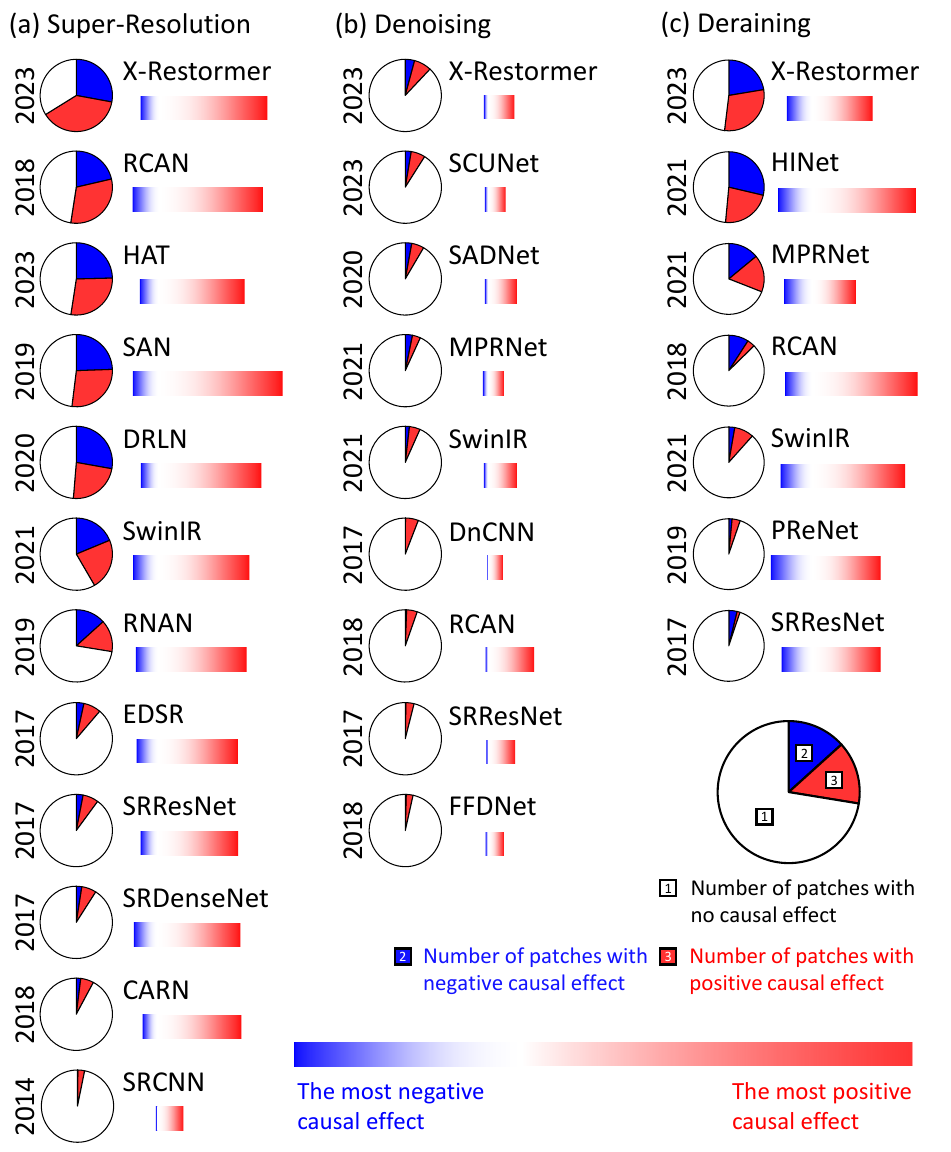}

    \caption{Average percentage of patches with positive and negative causal effects of all the models' CEMs on the test set is exhibited in the pie chart. The colorbars indicate the average causal effect range of models, which are scaled consistently. The proposal years of the models are also recorded.}
    \label{fig:summary}

\end{figure}

\begin{table}[t]
    \centering
    \caption{The quantitative CEM results of SR, DN, and DR models.}
    \setlength{\tabcolsep}{3pt}
    \label{tab:all}
    \begin{tabular}{lcccc}
        \toprule 
        Models &Positive (\%)  & Negative (\%) & None (\%) & Range (dB) \\ 
        \midrule
        \multicolumn{5}{c}{\textbf{Super-Resolution (SR)}} \\ 
        X-Restormer & 38.27 & 27.89 & 33.84 & (-0.21, 1.22) \\
        RCAN & 31.07 & 21.38 & 47.55 & (-0.31, 1.17) \\
         HAT & 27.77 & 24.64 & 47.59 & (-0.21, 0.97) \\
        SAN & 27.51 & 24.44 & 48.05 & (-0.29, 1.40) \\
        DRLN & 23.52 & 27.79  & 48.69 & (-0.21, 1.16) \\
        SwinIR & 22.85 & 18.75 & 58.40 & (-0.28, 1.04) \\
        RNAN & 14.25 & 13.32 & 72.43 & (-0.26, 1.01) \\
        EDSR & 7.85 & 3.41 & 88.74 & (-0.24, 0.91) \\
        SRResNet & 7.04 & 3.03 & 89.93 & (-0.19, 0.91) \\
        SRDenseNet & 6.59 & 2.45 & 90.96 & (-0.26, 0.94) \\
        CARN & 5.87 & 1.97 & 92.16 & (-0.16, 0.95) \\
        SRCNN & 3.06 & 0.12 & 96.81 & (-0.02, 0.30) \\
        \midrule 
        \multicolumn{5}{c}{\textbf{Denoising (DN)}} \\ 
        X-Restormer & 7.87 & 4.17 & 87.96 & (-0.05, 0.30) \\
        SCUNet & 6.36 & 2.63 & 91.01 & (-0.04, 0.20) \\
        SADNet & 5.59 & 2.87 & 91.54 & (-0.04, 0.33) \\
        MPRNet & 3.66 & 3.18 & 93.16 & (-0.06, 0.18) \\
        SwinIR & 4.85 & 1.83 & 93.32 & (-0.05, 0.33) \\
        DnCNN & 5.80 & 0.08 & 94.12 & (-0.01, 0.16) \\
        RCAN & 4.79 & 0.59 & 94.62 & (-0.03, 0.53) \\
        SRResNet & 3.65 & 0.28 & 96.07 & (-0.02, 0.31) \\
        FFDNet & 3.04 & 0.43 & 96.53 & (-0.03, 0.18) \\
        \midrule 
        \multicolumn{5}{c}{\textbf{Deraining (DR)}} \\ 
        X-Restormer & 29.55 & 22.38 & 48.07 & (-0.29, 0.68) \\
        HINet & 22.83 & 28.67 & 48.50 & (-0.38, 1.17) \\
        MPRNet & 16.94 & 14.02 & 69.04 & (-0.32, 0.49) \\
        RCAN & 3.55 & 9.13 & 87.32 & (-0.31, 1.19) \\
        SwinIR & 8.91 & 2.71 & 88.38 & (-0.36, 1.05) \\
        PReNet & 3.77 & 1.50 & 94.73 & (-0.47, 0.77) \\
        SRResNet & 1.35 & 3.80 & 94.85 & (-0.35, 0.77) \\
        \bottomrule 
    \end{tabular}
\end{table}
First, we compare the SR and DN networks. 
However, from the result in Fig. \ref{fig:summary}(b), the DN networks only focus on a small range of input information on average, regardless of the mechanism used by the network model to increase the receptive field.
There is only minor variation among different DN networks in the ability to collect input information.
Recalling the acquisition of CEM, we performed the intervention by only changing the image content without allowing the intervened image to deviate from its degradation distribution.
DN networks are not very responsive to image content outside of the ROI, where their focus is on the noise.

\begin{figure}[t]
    \centering
    \includegraphics[width=\linewidth]{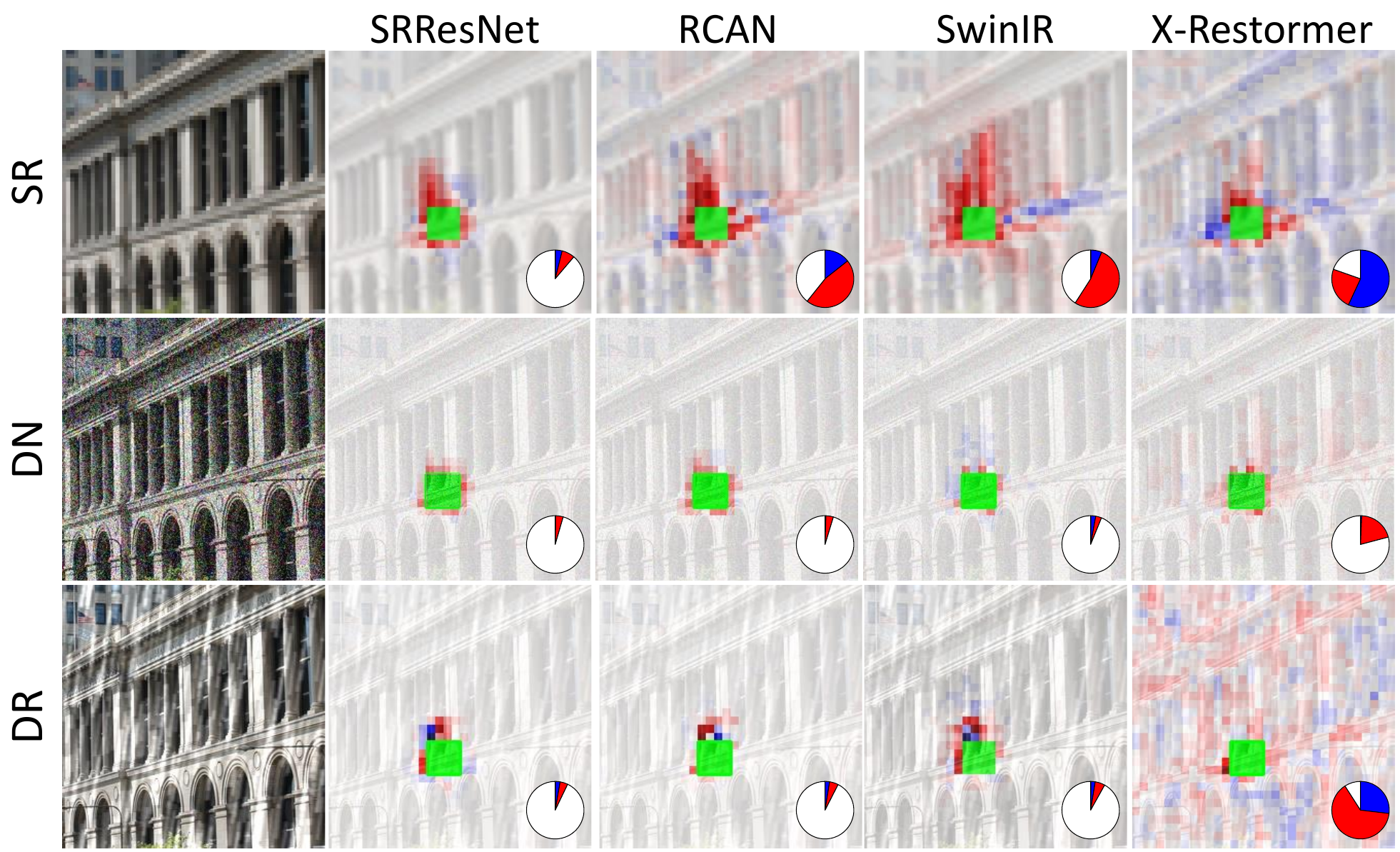}
    \caption{CEMs of four representative networks when dealing with different degradations of the same image.}
    \label{fig:SRDNDR}

\end{figure}

Rainy images in the experimental scenarios are generated by adding simulated rain streaks in the image. 
The rain streaks can be regarded as a certain pattern of streak noise.
Both noise and rain streaks belong to additive distortions and are independent of the image content.
Do DR networks show similar behavior as DN networks?
Results recorded in Fig. \ref{fig:summary}(c) show that models utilize more input information when solving the DR task than when dealing with the DN task.
The difference between rain and noise is that rain is regional, whereas the Gaussian noise is totally spatially independent.
It is feasible to enhance the information-gathering capacity of DR networks through well-designed mechanisms, but at the same time, there is no guarantee that all of this information will be positive.
CEMs of networks handling different LV tasks are presented in Fig. \ref{fig:SRDNDR}.
We can clearly see the difference in network behaviors.
\textbf{\textit{Regardless of the network architecture, the networks tend to focus on the ROI and neighboring pixels when dealing with the DN task.
Whereas, a well-designed network can take more patches into account when dealing with SR and DR tasks.}}
All the quantitative CEM results are recorded in Tab. \ref{tab:all}. 
The percentages of patches with positive, negative, and no causal effect on ROI, and the causal effect ranges of all the SR, DN, and DR models involved in our experiments are reported.

\subsection{Interpreting General LV Models}
In the previous single-task experiments, we forced the network to focus on one specific degradation and thus observed their behavioral preferences for single tasks.
We can compare the pie charts in Fig. \ref{fig:summary} of SRResNet, RCAN, SwinIR, and X-Restormer in all three tasks.
It can be observed that the SR task motivates networks with a global respective field to utilize more information.
In contrast, the network dealing with DN can address the task well by considering only a smaller range of input patches, even though the network has the potential to capture more input information.
The current trend in research is pursuing the generality of models. 
In the field of LV, this implies a model's capability to reconstruct various degraded images.
Naturally, we are curious about the following question: \textit{\textbf{Do general models trained with the mixed training setting behave consistently with their single-task version?}}

We consider handling SR, DN, and DR tasks jointly in this work.
The training dataset consists of low-resolution, noisy, and rainy images as the low-quality input images.
All of the training samples are obtained from DIV2K via the simulation process mentioned in Sec. \ref{setting}.
The sample probability for each task is the same.
Three representative networks, namely SRResNet, RCAN, and SwinIR are trained in this multiple-in-one setting.
They represent three widely adopted structures in network design, \ie residual block, attention mechanism, and Transformer respectively.

\begin{figure}[t]
    \centering
    \includegraphics[width=\linewidth]{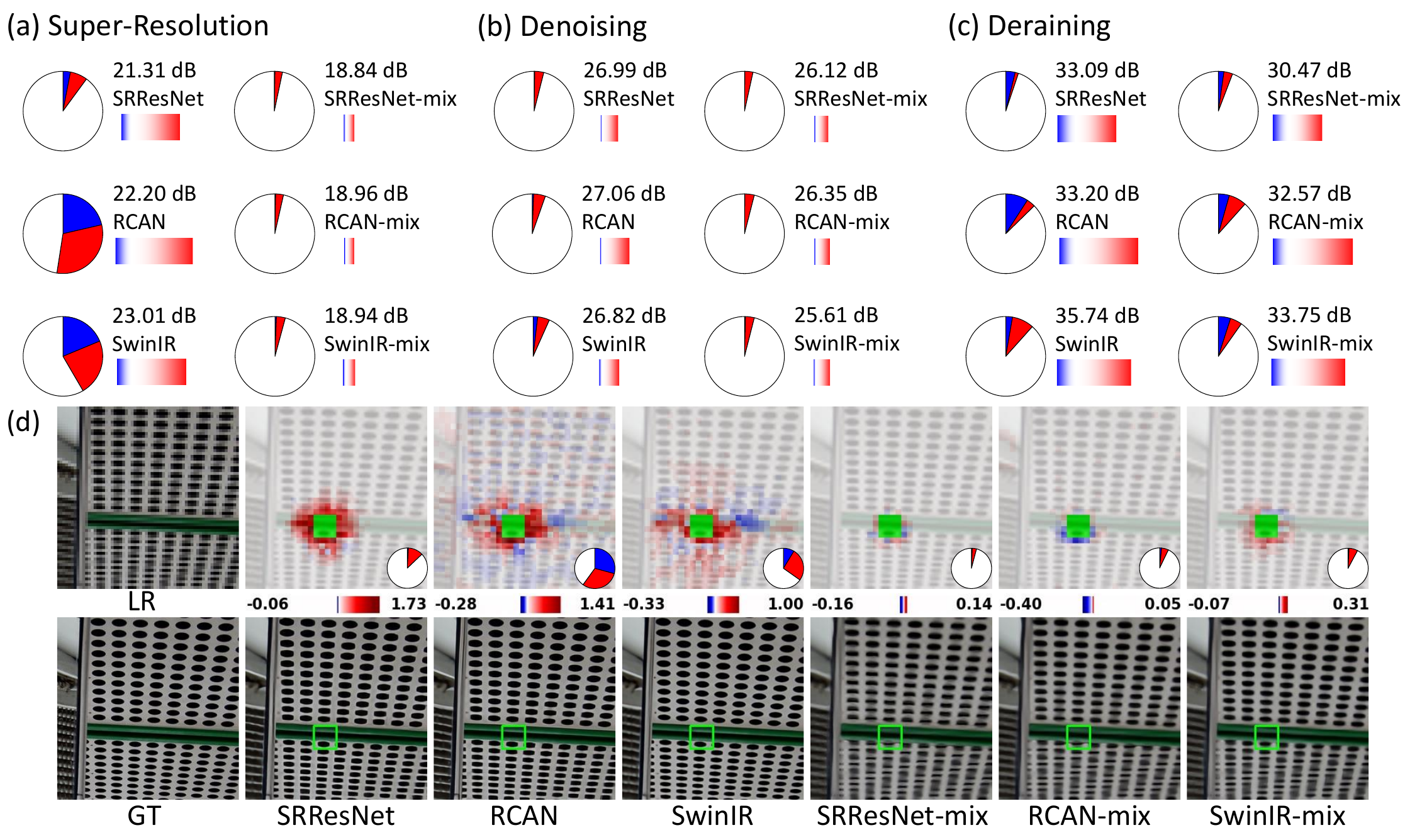}
    \caption{Comparison between the original networks and the versions using mixed training in (a) SR, (b) DN, and (c) DR. (d) An example of the behavior difference of networks with and without mixing training in SR.}

    \label{fig:mix}
\end{figure}

Fig. \ref{fig:mix}(a) exhibits the comparison of the original networks and their mix-trained versions.
It can be seen that the behavior of general networks when dealing with the SR task has changed dramatically compared to the previous single-task version.
Analyzing the statistics of CEMs in the SR task, we find that there are significantly fewer patches in the input image that have a causal relationship with ROI.
The range of causal effects that can be caused is also dramatically reduced.
Even networks with global receptive fields like RCAN and SwinIR lose the ability to utilize global information when dealing with SR.
The example in Fig. \ref{fig:mix}(b) shows the apparently different behavior of those networks.
All the networks trained with mixed tasks are degraded to only focus on neighbor areas of ROI.
On the other hand, the behaviors of networks dealing with DN and DR remained the same as their single-task version. 
The average PSNR values of models testing on the test set are also noted in Fig. \ref{fig:mix}.
The performance degradation of SR is the largest, while the performance drop of DN and DR is acceptable.
It indicates that networks are focusing on addressing the task of elimination, \ie DN and DR, but ignoring the task of generating details, \ie SR.
The network tends to utilize relatively local regions when dealing with DN and DR tasks, even if the model possesses a great ability to capture global information which is proven in their single-task version of image super-resolution.

Our method demonstrates the characteristic that SR is easily affected by DN and DR during the mixed degradation training. 
This specific aspect has not been explicitly addressed in previous low-level vision works. 
We use this example to demonstrate the value of our CEM, which provides researchers with an interpretability tool to analyze networks and tasks from perspectives beyond network performance.

\begin{figure}[t]
    \centering
    \includegraphics[width=\linewidth]{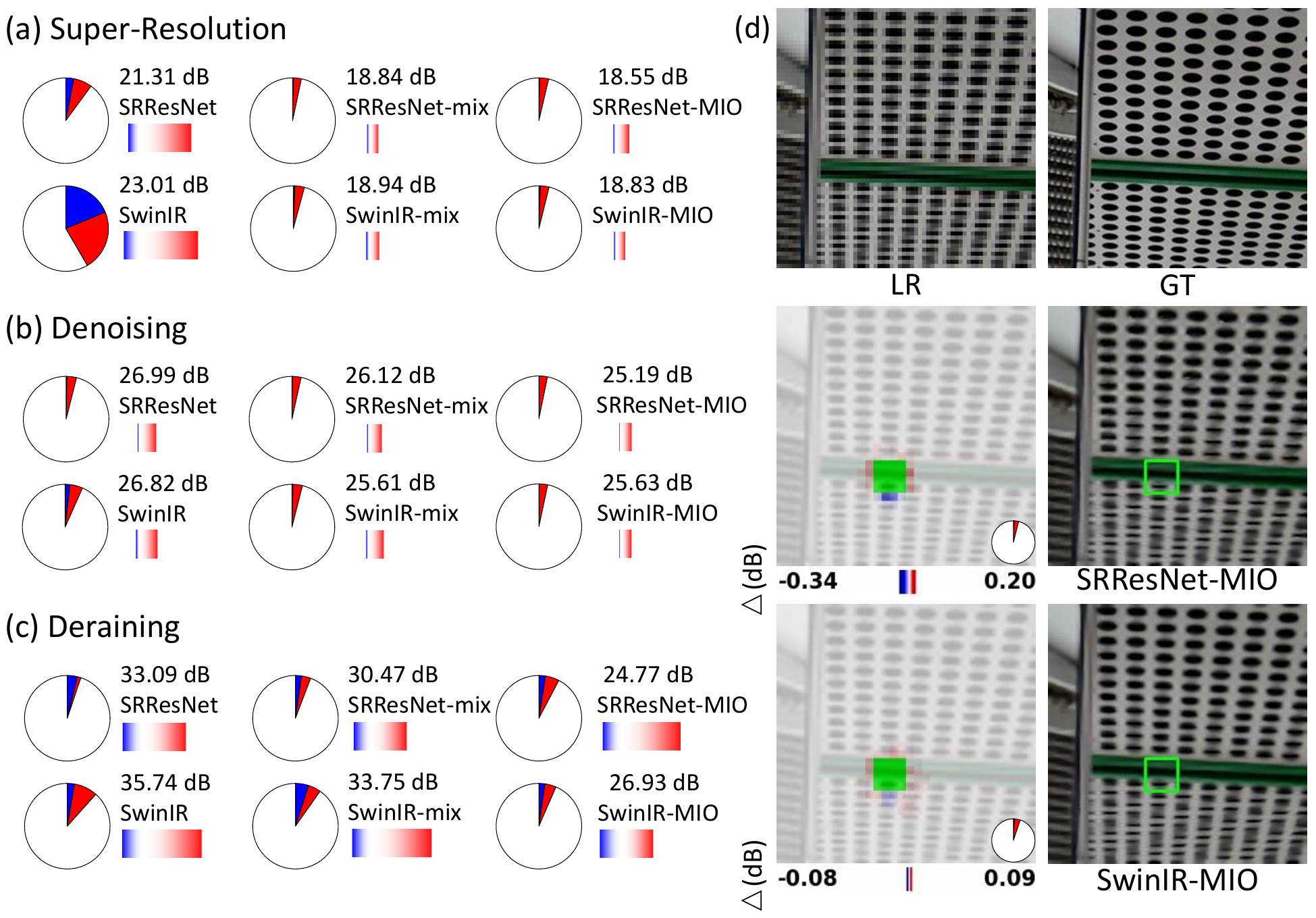}
    \caption{Comparison of the original networks with their mix and MIO versions in (a) SR, (b) DN, and (c) DR. (d) Example showing the behavior of SRResNet-MIO and SwinIR-MIO in SR.}
    \label{fig:MIO}
\end{figure}

Furthermore, We also conducted experimental analyses on pre-trained Multiple-In-One (MIO) networks trained with a broader range of tasks \cite{kong2024towards}. Specifically, SRResNet-MIO and SwinIR-MIO were trained with 7 equally mixed tasks, including super-resolution, deblurring, denoising, deJPEG, deraining, dehazing, and low-light enhancement. 
The CEM results are presented in Fig. \ref{fig:MIO}, which demonstrate the phenomenon that the MIO models utilize a more localized image area when dealing with SR task.

\textbf{\textit{When training a general LV model on a variety of tasks, the model tends to focus more on local knowledge, rather than leveraging its capacity to process global information for SR.}}
In other words, the network reverts to concentrating solely on local information, even though it is initially designed with the capability to gather and utilize globally.
Our analysis through CEM has revealed a critical focus for the future development of general low-level models: it is imperative to prevent the model from defaulting to a state that overemphasizes local, neighboring information, thereby losing its ability to incorporate long-range knowledge.

\section{Ablation Study}\label{ablation}
In this section, we conduct a comprehensive ablation study to evaluate the impact of various factors of our CEM. 
We systematically investigate the role of different components by testing on the natural image library, implementing different accelerating strategies, and examining the effects of varying patch sizes. 
This analysis helps identify the rationality and effectiveness of each component in our method. 
It is important to note that the super-resolution (SR) task typically includes a greater number of patches exhibiting causal effects, thereby more distinctly emphasizing the differences in CEM compared to other tasks.
As a result, our ablation study focuses solely on the domain of image SR.

\subsection{Natural Image Library}
\label{A-dataset}
\begin{figure}[t]
    \centering
    \includegraphics[width=\linewidth]{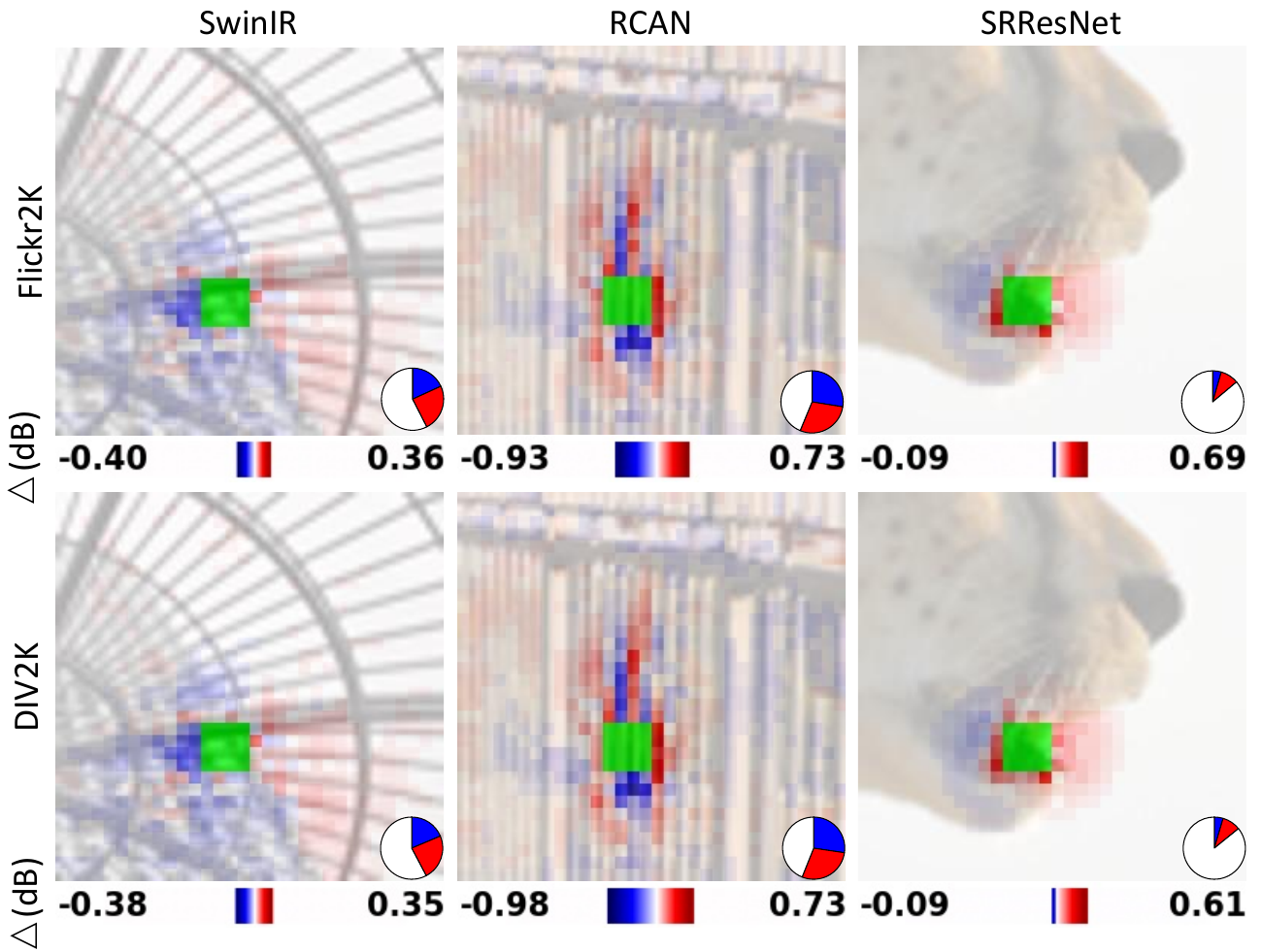}
    \caption{Different CEMs of SwinIR, RCAN, and SRResNet generated with patches sampled from DIV2K and Flickr2K.}
    \label{fig:A-dataset}
\end{figure}

The dataset mainly provides diverse samples from natural images as the intervention patches. 
As long as the dataset can represent a natural image distribution, the intervention can be conducted effectively. 
To demonstrate it, we perform the same CEM calculations with different datasets. 
We compare all the CEMs of the test images of three representative SR methods (SwinIR, RCAN, and SRResNet) generated by DIV2K and another natural image dataset Flickr2K \cite{timofte2017ntire}.
Tab. \ref{tab:A-dataset} displays the percentages of different causal effect patches and the effect ranges of those models.
It's evident that the choice of libraries does not result in significantly different outcomes. 
This indicates that our designed intervention method does not severely disrupt the distribution of images, nor does it significantly influence the behavior of the networks.
Furthermore, some qualitative examples illustrating the similarity are provided in Fig. \ref{fig:A-dataset}.
The examples presented show that the CEMs obtained through different datasets are also similar.
In other words, our designed interventions can yield stable CEM.
\begin{table}[t]
\centering
\caption{The quantitative CEM results of SwinIR, RCAN, and SRResNet generated with DIV2K and Flickr2K.}
\setlength{\tabcolsep}{2pt}
\begin{tabular}{llcccc}
\toprule 
Datasets & Models & Positive (\%)  & Negative (\%) & None (\%) & Range (dB) \\ 
\midrule 
\multirow{3}{*}{DIV2K} & SwinIR &22.85  & 18.75  & 58.40  & (-0.28, 1.04) \\ 
 & RCAN & 31.07  & 21.38  & 47.55  & (-0.31, 1.17) \\ 
 & SRResNet & 7.04  & 3.03  & 89.93  & (-0.19, 0.91) \\ 
 \midrule 
\multirow{3}{*}{Flickr2K} & SwinIR & 23.24  & 19.89  & 56.87  & (-0.30, 1.01) \\ 
 & RCAN & 30.85  & 21.14  & 48.01  & (-0.31, 1.14) \\ 
 & SRResNet & 7.05  & 3.04  & 89.91  & (-0.18, 0.96) \\ 
 \bottomrule
\\ 
\end{tabular}
\label{tab:A-dataset}
\end{table}

\begin{figure}[t]
    \centering
    \includegraphics[width=\linewidth]{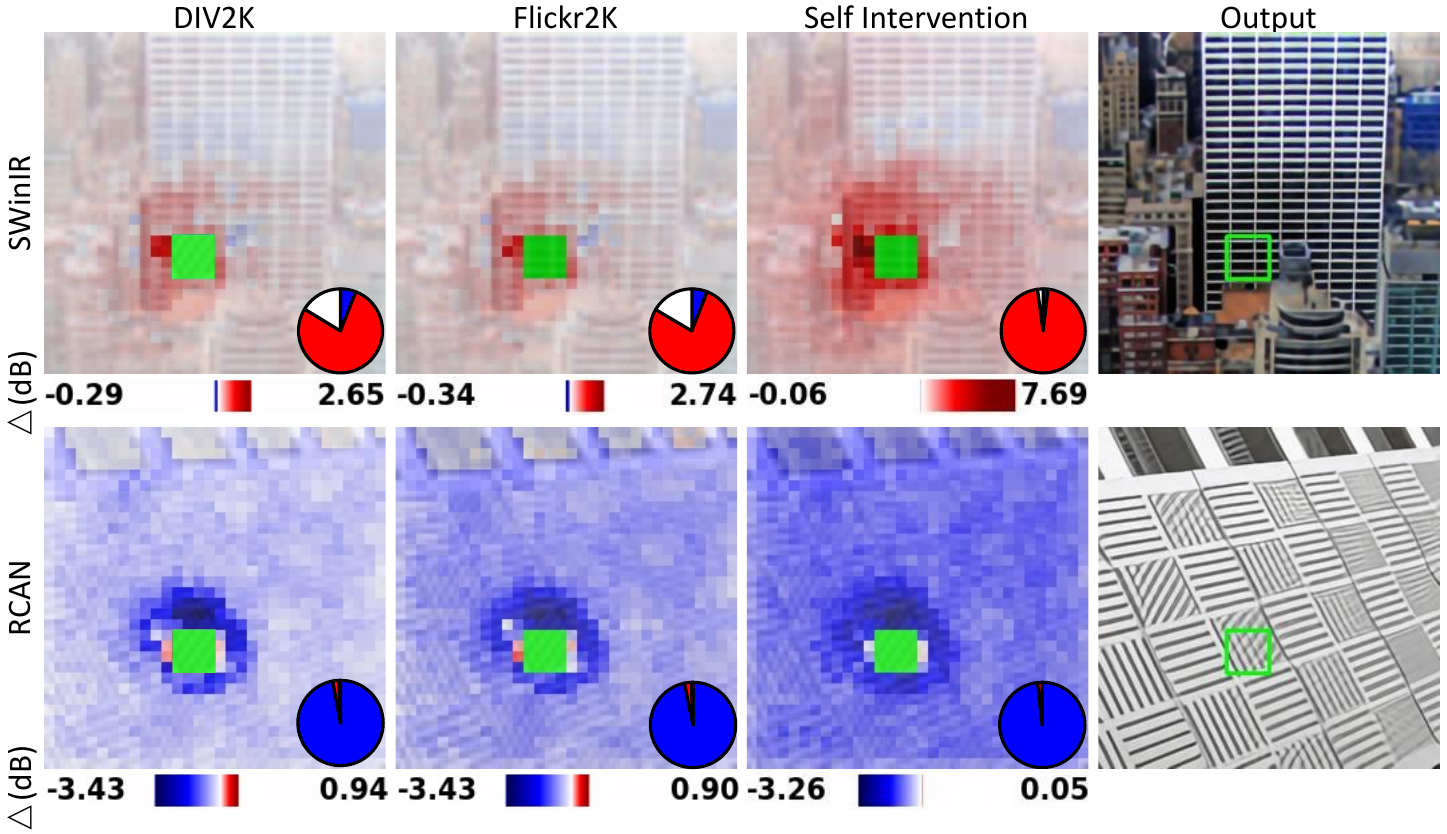}
    \caption{Different CEMs of SwinIR, RCAN generated with different intervention datasets.}
    \label{fig:A-self}
\end{figure}

Additionally, cropping patches directly from the input image itself appears to be an option, which ensures that the patch content does not shift dramatically.
However, there is a possibility that cropping patches from the input image may introduce biases limited to just a specific scene, potentially leading to less representative results.
To illustrate this point, we can observe the simple example in the comparison in Fig. \ref{fig:A-self}, the experiment demonstrates that selecting patches from the input image itself is not advisable.
CEMs generated from DIV2K and Flickr2K exhibit similarity, whereas CEMs generated from selecting patches from the input image itself are completely different from those from DIV2K and Flickr2K.
Therefore, we have chosen to utilize patches from a natural image library to ensure that the intervention patches faithfully represent the entire natural image distribution.
This decision aims to minimize biases and ensure the robustness and generalizability of the intervention method.

\begin{figure}[t]
    \centering
    \includegraphics[width=\linewidth]{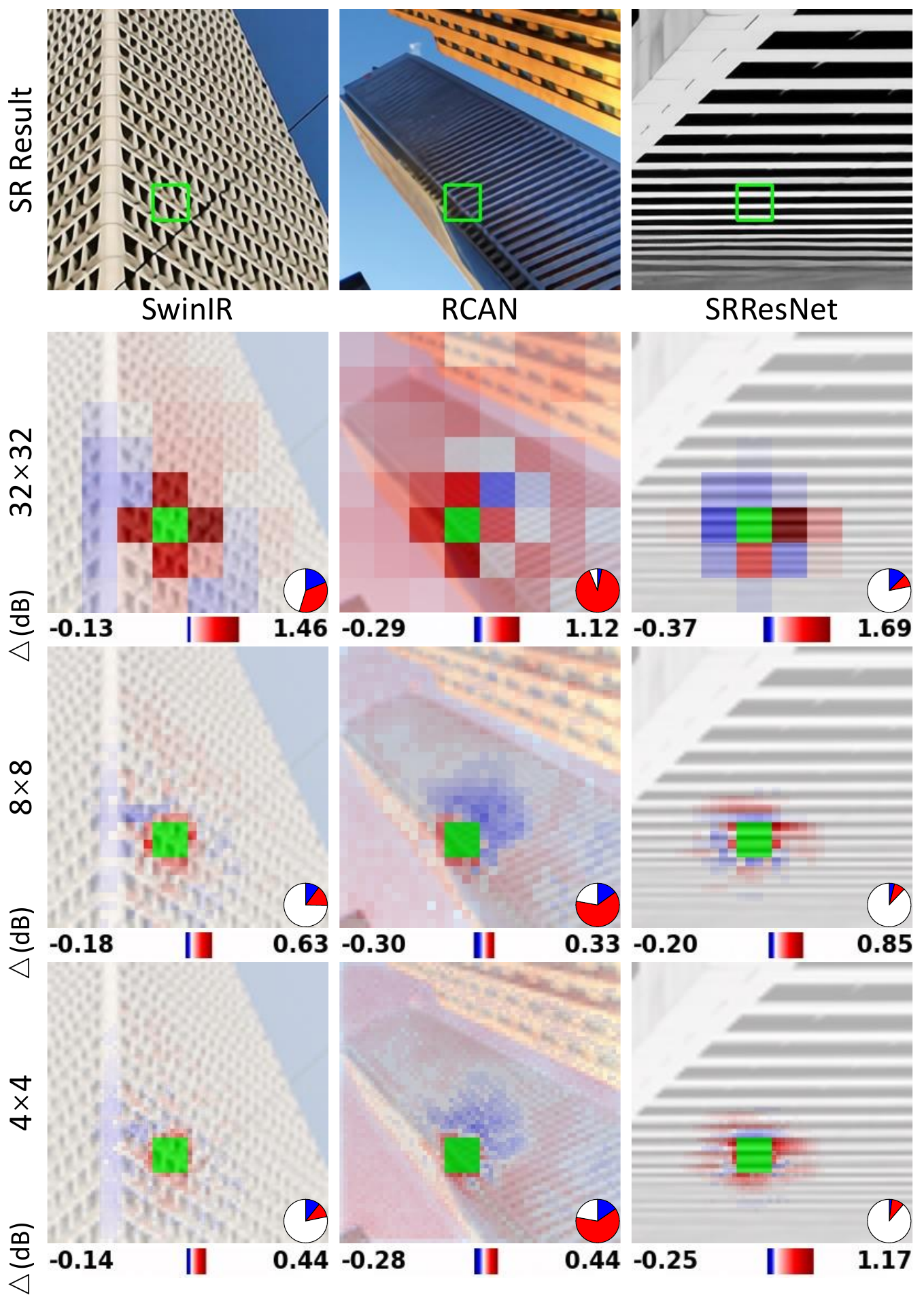}
    \caption{Different CEMs of SwinIR, RCAN, and SRResNet generated with different patch sizes.}
    \label{fig:A-size}
\end{figure}

\subsection{Patch Size}
\label{A-size}
Regarding the intervention patch size, it can be regarded as a hyperparameter, as different patch sizes lead to varying resolutions of CEM and affect the inference times.
Additionally, the patch size entails a trade-off. 
A larger size reduces computational load but may induce more harmful perturbation, whereas a smaller size increases computational time and may result in less meaningful interventions with less harm.

For an \( N \times N \) image:
(i) With a patch size of \( 32 \times 32 \), the image will be divided into \(\left(\frac{N}{32}\right) \times \left(\frac{N}{32}\right)\) patches.
(ii) With a patch size of \( 8 \times 8 \), the image will be divided into \(\left(\frac{N}{8}\right) \times \left(\frac{N}{8}\right)\) patches, which is 16 times more patches than the \( 32 \times 32 \) size.
(ii) With a patch size of \( 4 \times 4 \), the image will be divided into \(\left(\frac{N}{4}\right) \times \left(\frac{N}{4}\right)\) patches, which is 64 times more patches than the \( 32 \times 32 \) size.
We recommend dividing the image into \( 32 \times 32 \) patches, which corresponds to a patch size of \( 8 \times 8 \) for an image of \( 256 \times 256 \). 
This approach, as used in our paper, achieves a proper balance between computational time and CEM accuracy.
Some CEM examples with different intervention patch sizes can be found in Fig. \ref{fig:A-size}. 
Overall, the general shapes of these CEMs are similar, yet the inference times may vary significantly.

Consequently, various natural image datasets tend to yield similar results, and different patch sizes reflect similar overall shapes.
Furthermore, LAM \cite{gu2021interpreting} employs an integral gradient method to obtain correlation maps between input and output without modifying the networks and images.
We can also observe that our CEM results exhibit highly similar shapes to LAM results, as illustrated in Fig. \ref{fig:corr2} and Fig. \ref{fig:LAMvsCEM}.
This consistency offers additional evidence supporting the validity and efficacy of our CEM approach.

\subsection{Accelrating Strategy}
\label{A-accel}
As mentioned in Sec. \ref{ss:accelrating}, the intervention patches are sampled from the natural image library according to the probability density $\mathcal{D}$.
We now analyze the effectiveness of the probability density $\mathcal{D}$, which is derived from image gradients. The image gradient of an image \(I\) is calculated as follows:
\begin{equation}
\small
    G = \sqrt{\|(I(x,:)-I(x-1,:)\|_2^2+\|(I(:,y)-I(:,y-1)\|_2^2},
\end{equation}
where \(x\) and \(y\) denote the indices of columns and rows, respectively. $\| \cdot \|_2$ is the $\ell_2$ norm. 
\begin{figure}[t]
    \centering
    \includegraphics[width=\linewidth]{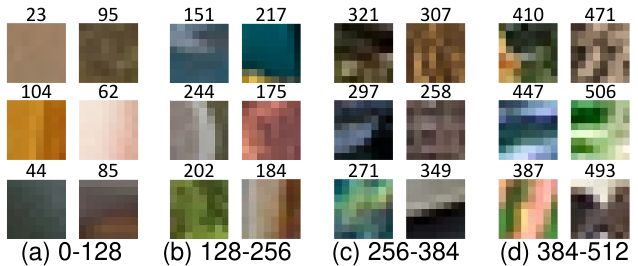}

    \caption{Several visual examples of patches with \( G \) values sampled from the following ranges are presented: 
(a) 0--128, (b) 128--256, (c) 256--384, and (d) 384--512. 
The corresponding \( G \) values are indicated above each patch.}
    \label{fig:sharpeness_examples}

\end{figure}

Several samples from different ranges of $G$ are presented in Fig. \ref{fig:sharpeness_examples}.
We observe that patches with smaller $G$ values tend to have simpler and smoother content, while patches with larger $G$ values exhibit more complex structures and greater pixel-to-pixel differences.
Additionally, in the probability density $\mathcal{{D}}$, patches sampled with smaller $G$ values are more frequent, which aligns with intuition—flat regions are more prevalent than regions with significant variations in natural images.
It should be noted that the method of calculating the gradient does not need to strictly adhere to the above equation, as long as it serves as an effective measure of variation in the content of the image.
Additionally, the hyperparameters for the number of interventions in the coarse and fine stages of CEM calculation can be discussed.

The proportions of inference times and similarity scores for different intervention strategies compared to the primary pipeline of CEM calculation (Eq. \ref{eq:500}) are recorded in Tab. \ref{tab:A1}.
It shows the effectiveness of different intervention strategies, where C and F denote the inference times in the coarse and fine stages, respectively, and U represents uniform sampling. 
Our proposed method achieves the highest similarity score of 82.62\% with an inference time of 4.9\%.
Compared to "C3F50-U," which also has an inference time of 4.9\% but a lower similarity score of 77.72\%, our method demonstrates the superiority of probability density $\mathcal{D}$ for the natural image distribution. 
The "C1F50" strategy, with fewer coarse-stage inferences, achieves a similarity score of 75.86\% with a shorter inference time of 2.9\%, indicating the importance of balancing coarse and fine-stage interventions.
Our method balances the number of interventions to maximize similarity scores while maintaining efficient inference.

\begin{table}[t]

    \caption{The proportion of inference times and similarity scores for different intervention strategies. C and F denote the inference times used in the coarse stage and fine stage, respectively. U represents sampling with uniform probability.}\label{tab:A1}
\centering\renewcommand\arraystretch{1}\setlength{\tabcolsep}{6pt}
    \begin{tabular}{lcc}
                \toprule
                Strategies & Inferences (\%) & Similarity Score (\%) \\
                \midrule
                C3F10&1.3 & 65,32  \\
                C3F30&2.8  & 78.68  \\
                C3F50-U& 4.9 &77.72  \\
                C1F50& 2.9 &75.86  \\
                C3F50 (ours)& 4.9  &82.62  \\
                \bottomrule
                \\
  \end{tabular}
\end{table}

\section{Conclusion}
In this paper, we propose a general method for interpreting low-level vision (LV) models called CEM, which is model- and task-agnostic.
Besides, It indicates the positive and negative causal effect of input patches, which brings causality to the low-level vision interpretability for the first time.
By applying this tool to different LV models, We obtain some conclusions.
Network capturing a large range of the input image doesn't always have positive outcomes, and can even increase the risk of being misled.
Moreover, well-designed mechanisms with a global receptive field are ineffective when dealing with image denoising, as the network tends to focus solely on local areas.
By combining multiple low-level vision tasks to train a general model, we may end up with underperformance and a reliance on local information, even though the network with single-task training indeed has the capacity to gather global information.

\bibliography{Main}
\bibliographystyle{IEEEtran}

\begin{IEEEbiography}[{\includegraphics[width=1in,height=1.25in,clip,keepaspectratio]{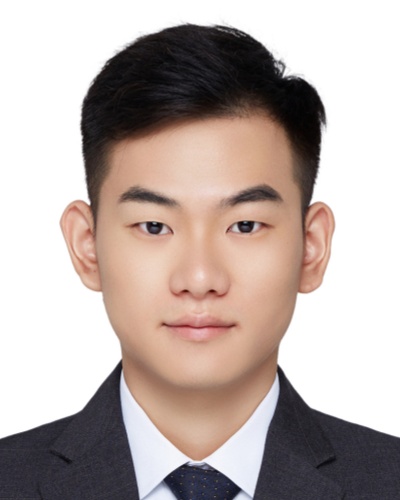}}]{Jinfan Hu}
is currently a Ph.D. student at Shenzhen Institutes of Advanced Technology, Chinese Academy of Sciences, under the supervision of Prof. Chao Dong. He received his B.S. and M.S. degrees from the University of Electronic Science and Technology of China, Chengdu, in 2019 and 2022, respectively. His research interests include low-level computer vision and the interpretability of deep learning.
\end{IEEEbiography}


\begin{IEEEbiography}[{\includegraphics[width=1in,height=1.25in,clip,keepaspectratio]{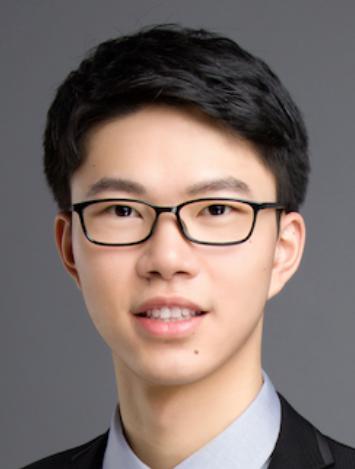}}]{Jinjin Gu} currently works at the Shanghai Artificial Intelligence Laboratory. He received his Ph.D. from the School of Electrical and Computer Engineering at the University of Sydney in 2024, under the supervision of Prof. Wanli Ouyang and Prof. Luping Zhou. Prior to that, he earned a Bachelor's degree in Computer Science and Engineering from The Chinese University of Hong Kong, Shenzhen, in 2020. He collaborates closely with Prof. Chao Dong and Prof. Junhua Zhao. His research focuses on computer vision and image processing, with additional interests in the interpretability of deep learning algorithms and the industrial applications of machine learning.

\end{IEEEbiography}


\begin{IEEEbiography}[{\includegraphics[width=1in,height=1.25in,clip,keepaspectratio]{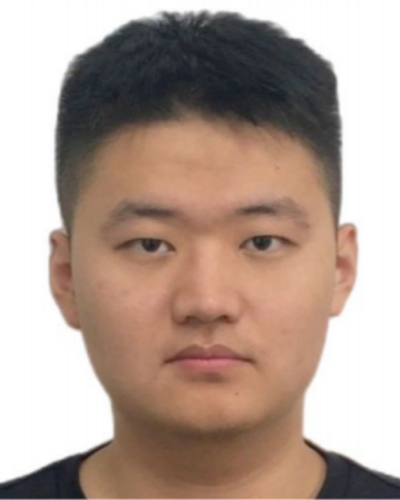}}]{Shiyao Yu} is currently a M.S. student in Southern University of Science and Technology and Shenzhen Institutes of Advanced Technology, Chinese Academy of Sciences. He is supervised by Prof. Chao Dong. He obtained his B.Eng. degree from Sichuan University, Chengdu, China. His research interests include image processing and deep learning.

\end{IEEEbiography}


\begin{IEEEbiographynophoto}{Fanghua Yu} currently works as a research intern in Multimedia Laboratory at Shenzhen Institutes of Advanced Technology, Chinese Academy of Sciences. He received the B.Eng. degree from Huazhong University of Science and Technology, Wuhan, China, in 2019, and the M.S. degree from Peking University, Beijing, China, in 2022. His research interests focus on image restoration.

\end{IEEEbiographynophoto}

\begin{IEEEbiography}[{\includegraphics[width=1in,height=1.25in,clip,keepaspectratio]{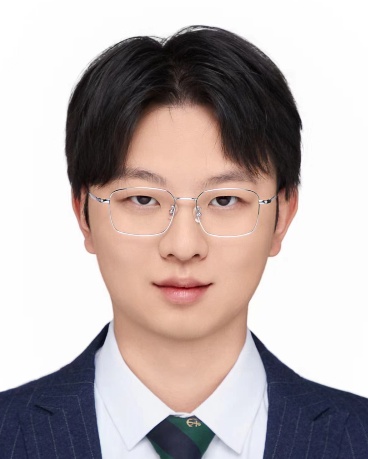}}]{Zheyuan Li} currently works as a research intern in Multimedia Laboratory at Shenzhen Institutes of Advanced Technology, Chinese Academy of Sciences. He received his B.S. degree from Northwestern Polytechnical University. He is supervised by Prof. Yu Qiao and Prof. Chao Dong. His research interests include computer vision and image super-resolution.

\end{IEEEbiography}


\begin{IEEEbiography}[{\includegraphics[width=1in,height=1.25in,clip,keepaspectratio]{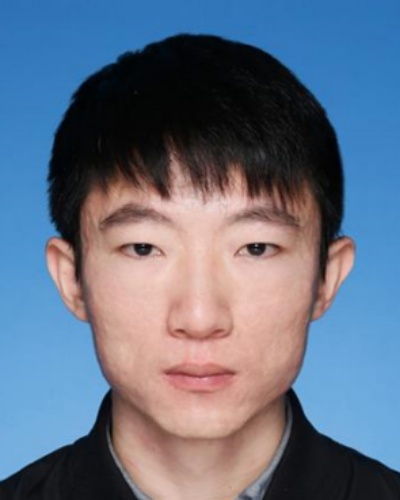}}]{Zhiyuan You} is a Ph.D. student from Multimedia Laboratory at The Chinese University of Hong Kong, under the supervision of Prof. Chao Dong and Prof. Tianfan Xue. He obtained both his Bachelor’s and Master’s degrees from Shanghai Jiao Tong University, where he was supervised by Prof. Xinyi Le and Prof. Yu Zheng. His research interests focus on low-level vision based on foundation models.

\end{IEEEbiography}

\begin{IEEEbiography}[{\includegraphics[width=1in,height=1.25in,clip,keepaspectratio]{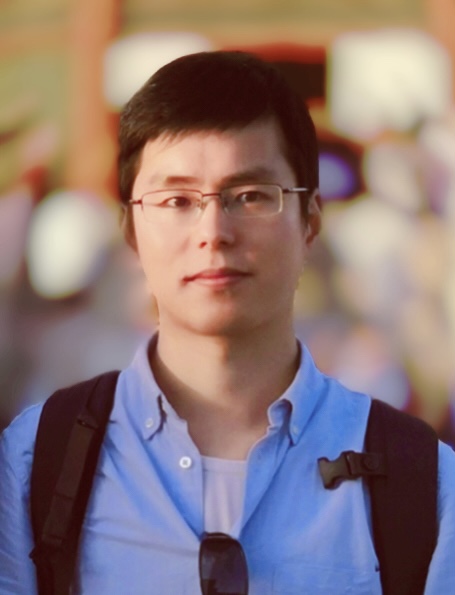}}]{Chaochao Lu} received the B.Eng. degree from Nanjing University, the M.Phil. degree from The Chinese University of Hong Kong, and the Ph.D. degree from the University of Cambridge. He currently works at the Shanghai Artificial Intelligence Laboratory. His research interests include causal machine learning and foundation models.

\end{IEEEbiography}

\begin{IEEEbiography}[{\includegraphics[width=1in,height=1.25in,clip,keepaspectratio]{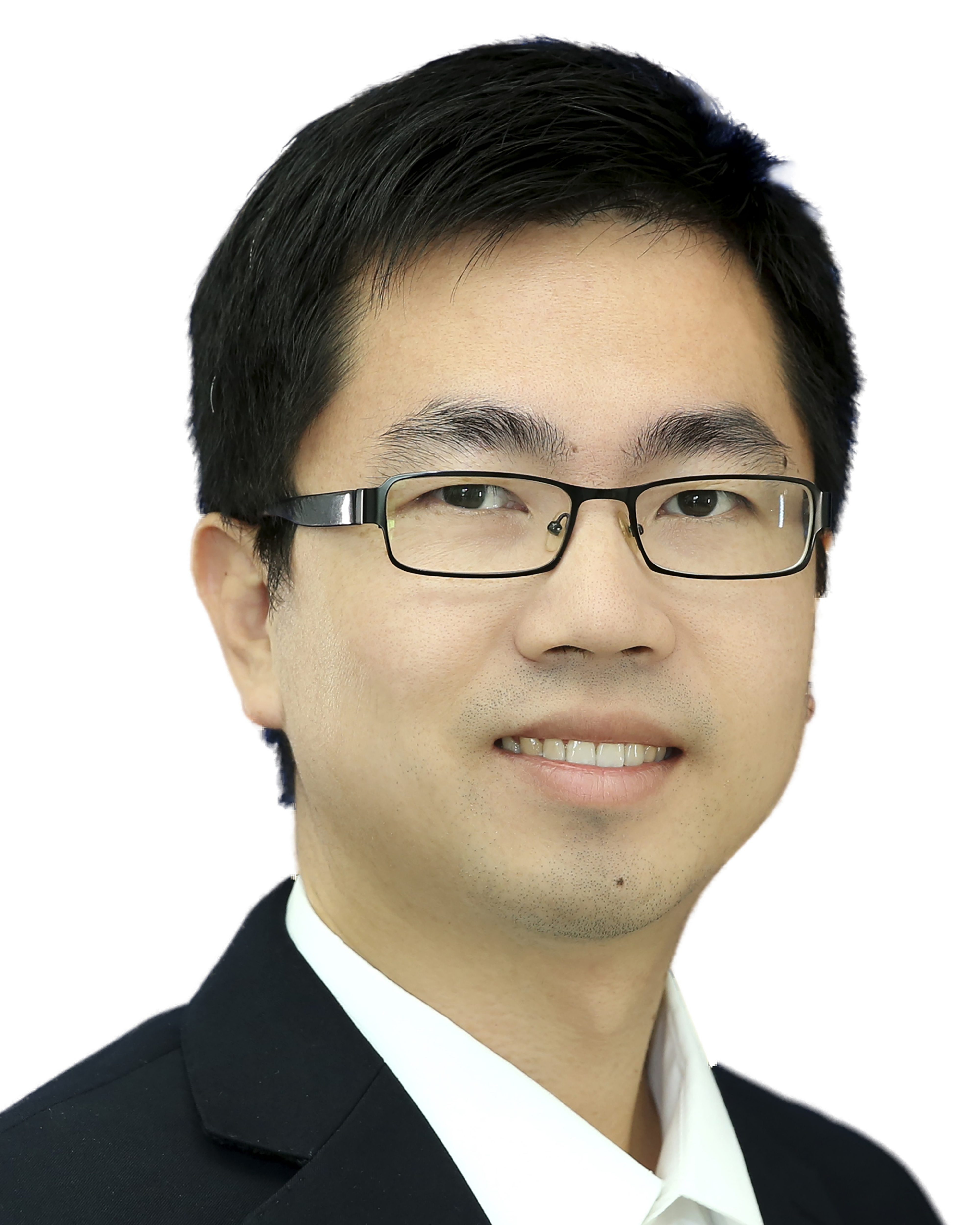}}]{Chao Dong} is a professor at Shenzhen Institutes of Advanced Technology, Chinese Academy of Science (SIAT), and Shanghai Artificial Intelligence Laboratory. In 2014, he first introduced deep learning method – SRCNN into the super-resolution field. This seminal work was chosen as one of the top ten ``Most Popular Articles" of TPAMI in 2016. His team has won several championships in international challenges –NTIRE2018, PIRM2018, NTIRE2019, NTIRE2020 AIM2020 and NTIRE2022. He worked in SenseTime from 2016 to 2018, as the team leader of Super-Resolution Group.  In 202l, he was chosen as one of the World's Top 2\% Scientists. In 2022, he was recognized as the Al 2000 Most Influential Scholar Honorable Mention in computer vision. His current research interest focuses on low-level vision problems, such as image/video super-resolution, denoising and enhancement.

\end{IEEEbiography}
\vfill

\end{document}